\documentclass[a4paper,11pt,oneside]{article}
\usepackage[absolute]{textpos}
\usepackage[T1]{fontenc}
\usepackage[utf8]{inputenc}

\usepackage{xspace}
\usepackage[table,xcdraw]{xcolor}
\usepackage{graphicx}
\usepackage{adjustbox}
\usepackage{subfig}
\usepackage{multirow}
\usepackage{amsmath,amssymb,amsfonts}
\usepackage{float}
\usepackage{hyperref}
\usepackage{url}
\usepackage{soul}
\usepackage{bm}
\usepackage[french,german,english]{babel}
\usepackage{utopia}
\usepackage[tracking=true,kerning=true,spacing=true]{microtype}
\usepackage[DIV=14,BCOR=2mm,headinclude=true,footinclude=false]{typearea}

\usepackage{setspace} 
\hypersetup{pdfborder={0 0 0}, colorlinks=true, linkcolor=black, citecolor=black, urlcolor=black}

\usepackage{biblatex}
\title{\textbf{Understanding Catastrophic Overfitting in \\ Adversarial Training}}

\author{ \hspace{1mm}\textbf{Peilin Kang}\thanks{The work is done at ETH Zurich as a master thesis.} \\
	\texttt{kangpeilin1995@gmail.com} \\ \\

    \hspace{1mm}\textbf{Seyed-Mohsen Moosavi-Dezfooli}\thanks{Thesis supervisor.} \\
	Institute for Machine Learning\\
	ETH Zurich\\
	Switzerland \\
	\texttt{seyed.moosavi@inf.ethz.ch} \\
}

\date{}

\begin{document}
  
\maketitle

\begin{abstract}
Recently, FGSM adversarial training is found to be able to train a robust model which is comparable to the one trained by PGD but an order of magnitude faster. However, there is a failure mode called catastrophic overfitting (CO) that the classifier loses its robustness suddenly during the training and hardly recovers by itself. In this paper, we find CO is not only limited to FGSM, but also happens in $\mbox{DF}^{\infty}$-1 adversarial training. Then, we analyze the geometric properties for both FGSM and $\mbox{DF}^{\infty}$-1 and find they have totally different decision boundaries after CO. For FGSM, a new decision boundary is generated along the direction of perturbation and makes the small perturbation more effective than the large one. While for $\mbox{DF}^{\infty}$-1, there is no new decision boundary generated along the direction of perturbation, instead the perturbation generated by $\mbox{DF}^{\infty}$-1 becomes smaller after CO and thus loses its effectiveness. We also experimentally analyze three hypotheses on potential factors causing CO. And then based on the empirical analysis, we modify the RS-FGSM by not projecting perturbation back to the $l_\infty$ ball. By this small modification, we could achieve $47.56 \pm 0.37\% $ PGD-50-10 accuracy on CIFAR10 with $\epsilon=8/255$ in contrast to $43.57 \pm 0.30\% $ by RS-FGSM and also further extend the working range of $\epsilon$ from 8/255 to 11/255 on CIFAR10 without CO occurring.
\end{abstract}

\tableofcontents

\section{Introduction}
In recent years, deep learning has achieved great success on many different tasks, such as image classification \cite{He_2016} and language modeling \cite{Devlin_2019}. However, those deep neural networks can be easily attacked by adding small, well-designed, imperceptible perturbations into clean examples \cite{Biggio_2013} \cite{szegedy2013intriguing}, which raises security concern when we deploy those models into real-world applications, such as autonomous driving. It would be devastating if a model misclassifies a stop sign to a straight-ahead sign because someone adds well-designed perturbations which are imperceptible by humans. Thus, it’s important to train a robust model which is not only accurate on normal clean examples but also on small perturbed adversarial examples. 

There are many different types of defense methods, such as preprocessing based methods \cite{guo2017countering}\cite{buckman2018thermometer}
\cite{song2017pixeldefend}\cite{samangouei2018defensegan}, regularization based methods \cite{ross2017improving}\cite{moosavidezfooli2018robustness}\cite{qin2019adversarial}, provable defenses on norm-bounded perturbation \cite{wong2017provable}\cite{raghunathan2018certified}\cite{gowal2018effectiveness}\cite{cohen2019certified} and adversarial training methods \cite{goodfellow2015explaining}\cite{tramèr2020ensemble}\cite{Kurakin_2018}\cite{madry2019deep}\cite{zhang2019theoretically}\cite{shafahi2019adversarial}. Some defenses are found to give a false sense of security because of gradient obfuscation, and they could be defeated by well-designed ad-hoc attacks \cite{athalye2018obfuscated}. Also, a recent paper \cite{croce2020reliable} evaluates top 50 defensive methods and finds most of them either lead to lower robust accuracy or will be broken using a stronger attack. In the end, adversarial training using PGD attack \cite{madry2019deep} or its variations \cite{zhang2019theoretically}\cite{mosbach2018logit}\cite{Xie_2019} leads to the most stable model which has the best empirical robustness when it faces different attacks. Thus, we will focus on adversarial training methods in this paper.

Adversarial training is to train models on adversarial examples constructed by different attack methods. Although we can get a robust model using projected gradient descent (PGD) adversarial training \cite{madry2019deep}, the main drawback of this method is heavy computational overhead resulted from using multi-step gradient descent to generate adversarial examples in each mini-batch weight updates. PGD adversarial training is an order of magnitude slower than standard training, and thus limits scalability to large neural networks, such as ImageNet. Though there are many works trying to accelerate PGD without having to sacrifice its performance, such as \cite{zhang2019theoretically} \cite{shafahi2019adversarial}\cite{zhang2019propagate}, they are still much slower than standard training.

In order to improve computational efficiency, fast gradient sign method (FGSM) adversarial training \cite{goodfellow2015explaining} uses one gradient step to construct adversarial examples instead of using multi-step gradient descent. This method is computationally efficient, but can be easily defeated by stronger multi-step attacks \cite{tramèr2020ensemble}\cite{kurakin2016adversarial}. However recently, \cite{wong2020fast} claims by simply adding random initialization before FGSM, we can get a model with comparable robustness as the one trained by multi-step methods, such as PGD. But this approach can not always defend against strong multi-step attacks, and this paper\cite{wong2020fast} presents a failure mode named \textbf{catastrophic overfitting (CO)} that the model suddenly losses robustness in a few epochs during training and barely recover by itself. Although the paper \cite{wong2020fast} finds that we can use a small validation set to evaluate the model during training and stop it before catastrophic overfitting, the model we get is still sub-optimal because of insufficient training. Although there are some works trying to understand why catastrophic overfitting happens and improve adversarial training \cite{andriushchenko2020understanding}\cite{li2020understanding}\cite{Vivek_2020}\cite{kim2020understanding}, none of them can fully explain the reason for catastrophic overfitting and the methods proposed in these papers add additional computational overhead to FGSM. 

In this project, we work on understanding catastrophic overfitting in adversarial training and try to improve it. First, we find catastrophic overfitting is not only limited to FGSM adversarial training, but also happens on RS-FGSM \cite{wong2020fast} and $\mbox{DF}^{\infty}$-1 \cite{moosavidezfooli2016deepfool} (DF's superscript of $p$ means using $p$ norm to calculate perturbation at each iteration, and the number after dash means how many iterations to use. If the number of iteration is not specified, it means to stop until finding an adversarial example). Then we analyze the geometric properties of the model before and after catastrophic overfitting. Specifically, we draw the cross-section of the decision boundary spanned by two vectors. One vector is calculated by DeepFool($\mbox{DF}^2$) \cite{moosavidezfooli2016deepfool} which is perpendicular to the decision boundary, and the other one is calculated by the adversarial method used in the training process. We observe that FGSM and $\mbox{DF}^{\infty}$-1 show different geometric properties after catastrophic overfitting. For FGSM, a new decision boundary is generated along the perturbed direction, and small FGSM perturbation becomes more effective than large FGSM perturbation which causes the perturbation found by full step FGSM to become invalid. While for $\mbox{DF}^{\infty}$-1, though large perturbation is still more effective than small perturbation, the length of the perturbation generated by $\mbox{DF}^{\infty}$-1 decreases and becomes invalid. These geometric properties explain why an adversary loses its effectiveness after catastrophic overfitting happens and hardly recovers by itself. Then we focus on analyzing factors that cause catastrophic overfitting. Specifically, we propose experiments to examine three probable hypotheses, two of them put forward by previous works and one of them come up by us. And based on the empirical analysis, we modify the RS-FGSM by not projecting perturbation back to the $l_\infty$ ball, in other words, not clipping the perturbation to the $[-\epsilon, \epsilon]^d$. By this small modification, we could achieve $47.56 \pm 0.37\% $ PGD-50-10 accuracy on CIFAR10 at $\epsilon=8/255$ in contrast to $43.57 \pm 0.30\% $  by RS-FGSM under the same setting and also further extend the working range of $\epsilon$ (radius of $l_\infty$ ball) from 8/255 to 11/255 on CIFAR10 without suffering from catastrophic overfitting. Our work makes the following \textbf{contributions}:
\begin{itemize}
    \item We show that catastrophic overfitting is a general phenomenon and not only limited to FGSM, but also happens in RS-FGSM and $\mbox{DF}^{\infty}$-1 adversarial training.
    \item We analyze the geometric properties of classifiers before and after catastrophic overfitting and demonstrate that FGSM and $\mbox{DF}^{\infty}$-1 have similar decision boundaries before catastrophic overfitting and become totally different after catastrophic overfitting happens.
    \item We experimentally analyze three hypotheses on potential factors causing catastrophic overfitting. 
    \item We make a modification to RS-FGSM \cite{wong2020fast} by not projecting perturbation back to the $l_\infty$ ball. On standard dataset CIFAR10, we show that this modification leads to a better robust accuracy and it permits us to use larger values of $\epsilon$.
\end{itemize}

In section 2, we will overview the CO problem in adversarial training and some related works trying to understand and solve it. In section 3, we define the formal definition of CO and show it happens both in FGSM and $\mbox{DF}^{\infty}$-1 adversarial training. Then we analyze the geometric properties before and after CO in section 4 and experimentally analyze three probable hypotheses on factors causing CO in section 5. Finally, In section 6, we conclude our work and propose some future work.

\section{Problem overview and related work}
In this section, we will introduce adversarial training in detail, especially one gradient step method FGSM. 
Then we introduce the catastrophic overfitting phenomenon proposed in \cite{wong2020fast} which causes previous attempts to train robust models with FGSM and its variants to fail. Finally, we review some related works which try to understand why catastrophic overfitting happens and how to prevent it from happening or recover the model once it happens.

\subsection{Adversarial Training}
 Adversarial training is one of the most effective defensive methods to train a robust model which is not only accurate on clean examples, but also on the corresponding adversarially perturbed examples \cite{athalye2018obfuscated}. Adversarial examples are generated by adding small, well-designed perturbation vectors to clean examples. The difference between adversarial and clean examples are imperceptible by humans but they are categorized into different classes by the classifier. Standard training is not enough to train a robust model to resist adversarial examples, this is why adversarial training comes in. The basic idea of adversarial training is to construct adversarial examples and then apply empirical risk minimization on those examples instead of the clean ones. Given a dataset $(x, y)\sim \mathcal{D}$, $\mathcal{D}$ is the underlying data distribution, a model $f$ with parameter $\theta$, a loss function $l$ and a threat model $\Delta$, adversarial training can be formulated as the following min-max problem\cite{madry2019deep}:
\begin{equation}
     \displaystyle\min_{\theta} \mathbb{E}_{(x,y)\sim\mathcal{D}}[\max_{\delta\in\Delta}l(f_{\theta}(x+\delta), y)]
\end{equation}
It is well studied on how to minimize outside empirical risk, such as SGD and Adam methods. So the main focus of recent works is on how to find $\delta^{*} \in \Delta$ of a given data point $x$ that maximizes inner loss function. In this project, we focus on $l_{\infty}$ bounded threat model $\Delta= \{\delta \mid \|\delta\|_{\infty} \leq \epsilon, \epsilon > 0\}$. The different adversaries we use to solve the inner maximization problem constitute different adversarial training methods. For simplicity, we will use $l(x+\delta, y; \theta)$ to represent $l(f_{\theta}(x+\delta), y)$ later in this report. 

\textbf{Projected Gradient Descent (PGD)} \cite{madry2019deep} is state-of-the-art defense method which has not been broken by different attack methods \cite{athalye2018obfuscated}\cite{croce2020reliable}. 
    \begin{equation}\label{PGD}
    \begin{aligned}
        \delta^0 &\sim \mathcal{U}([-\epsilon, \epsilon]^d) \\\
        \delta^{t+1}&=\textstyle\prod_{[-\epsilon, \epsilon]^d} (\delta^{t}+\alpha \mbox{sign}(\nabla_{x}l(x+\delta^{t},y; \theta)))
    \end{aligned}
    \end{equation}
As shown in Equation \ref{PGD}, PGD first selects a random start point $\delta^0$ inside the $l_\infty$-ball and then uses multi-steps to solve the inner maximization problem with step size $\alpha < \epsilon$ in each step. During each iteration, the perturbation will be projected back to the $l_\infty$-ball. In order to get better inner maximization quality, PGD also adds several random restarts inside the $l_\infty$-ball. 

The main drawback of PGD is the huge computational overhead, which can be an order of magnitude higher than standard training and limits its scalability to train the large deep neural networks, such as ImageNet. 

In order to reduce computational cost, there is another method called \textbf{Fast Gradient Sign Method (FGSM)} \cite{goodfellow2015explaining}.
FGSM uses one gradient step to approximate the inner maximization problem.
\begin{equation}\label{FGSM}
     \delta_{FGSM} = \alpha \mbox{sign}(\nabla_{x}l(x,y; \theta))
\end{equation}
As shown on Equation \ref{FGSM}, FGSM takes the sign of input gradient as the direction where loss increases most rapidly and then multiplies it with step size $\alpha$ (in paper\cite{goodfellow2015explaining} $\alpha=\epsilon$). However, \cite{tramèr2020ensemble}\cite{kurakin2016adversarial} claims FGSM is only workable under limited circumstances, for example under small $\epsilon$. When $\epsilon$ is large, the model trained by FGSM can be broken by stronger attacks, such as PGD.

Recently, the paper \cite{wong2020fast} revisits FGSM and comes up with a method named \textbf{Fast Gradient Sign Method with Random Initialization (RS-FGSM)}:
\begin{equation}\label{RS-FGSM}
\begin{aligned}
    \delta &\sim \mathcal{U}([-\epsilon, \epsilon]^d) \\\
    \delta_{RS-FGSM} &= \textstyle\prod_{[-\epsilon, \epsilon]^d}(\delta+\alpha \mbox{sign}(\nabla_{x}l(x+\delta,y;\theta)))
\end{aligned}
\end{equation}
The paper \cite{wong2020fast} claims that FGSM adversarial training combined with \textbf{random initialization} is as effective as PGD-based training but an order of magnitude faster. This paper \cite{wong2020fast} also demonstrates a failure pattern, named \textbf{catastrophic overfitting} (CO) and states this might be the reason why previous attempts at FGSM and its variations fail. Catastrophic overfitting is a phenomenon where robust accuracy with respect to a strong adversary (e.g., PGD) suddenly and drastically drops to 0\% after several training epochs and rarely recovers by itself.

This paper \cite{wong2020fast} does not explain why adding random initialization can make FGSM workable with $\epsilon=8/255$ on CIFAR10 while vanilla FGSM fails. And it also does not explain why RS-FGSM still suffers from catastrophic overfitting when we further increase the step size $\alpha$ and how to avoid it. In the next subsection, we will introduce some related works trying to understand and solve catastrophic overfitting problems.

\subsection{Related works}
In this subsection, we will review some related works trying to understand catastrophic overfitting problems and improve adversarial training. The paper \cite{Vivek_2020} does not use the term catastrophic overfitting, but points out the same failure mode when training a  model using FGSM. The author thinks this is because of overfitting to FGSM perturbations and empirically shows that by adding dropout layer after all non-linear layers and decreasing dropout probability during the training process, we can train a model using FGSM and attain the same robustness as the one trained by stronger attacks, such as PGD. However, the proposed method is evaluated on small $\epsilon$, so it is hard to tell whether this dropout schedule is still effective or not when $\epsilon$ is larger. Another concern is that by using dropout after each non-linear layer, we may face the problem of underfitting. This paper came out almost the same time as \cite{wong2020fast}.

In the paper \cite{andriushchenko2020understanding}, the author fixes the step size of vanilla FGSM and RS-FGSM to be $\epsilon$ and $1.25\epsilon$ separately, and varied $\epsilon$ of $l_\infty$ threat model from 1/255 to 16/255. The author finds vanilla FGSM experiences catastrophic overfitting when $\epsilon > 6/255$, while RS-FGSM starts to fail when $\epsilon > 9/255$. Then it claims that the main reason why RS-FGSM is more effective than vanilla FGSM under larger $\epsilon$ is that random initialization decreases the expected magnitude of the perturbation. Finally, this paper proposes a regularization method, named GradAlign, which prevents catastrophic overfitting by explicitly maximizing the gradient alignment inside perturbation set to improve the linear approximation quality. By using FGSM+GradAlign, we can prevent catastrophic overfitting but add additional computational cost. 

Another paper \cite{li2020understanding} claims the key success factor of RS-FGSM is the ability to recover from catastrophic overfitting subject to weak attacks. And then they propose a simple fix: switch to multi-step PGD once catastrophic overfitting is detected, and switch back to FGSM after the model recovers to normal. This method is simple but does not explain why multi-step PGD can help the model recover from catastrophic overfitting and also adds additional computational cost.

The paper \cite{kim2020understanding} is a concurrent work of this project. The paper hypothesizes that catastrophic overfitting is caused by \textbf{decision boundary distortion}. The author finds when catastrophic overfitting happens, there is a trend of decreasing of expected $l_1$ norm of PGD7 perturbations  $\mathbb{E}_{(x,y) \sim \mathcal{D}}[\|\delta_{PGD-7}\|_1]$, increasing of expected squared $l_{2}$ norm of the input gradients $\mathbb{E}_{(x,y) \sim \mathcal{D}}[\|\nabla_{x}l(x,y;\theta\|_{2}]$, and the smaller perturbation can fool the classifier whereas the classifier is robust against larger perturbations. Based on this phenomenon, the author comes up with an idea that the decision boundary is distorted after catastrophic overfitting. And then the paper suggests a simple method that can prevent decision boundary distortion by searching appropriate step size for each input example during adversarial training. But searching minimum scaling k causes additional computation overhead:
\begin{equation} \label{decision distortion}
    \begin{aligned}
    \delta &= \epsilon \mbox{sign}(\nabla_x l(x, y; \theta)) \\
    k^{*} &=  \displaystyle\min_{k \in [0,1]}[k | y \neq f_{\theta}(x+k\delta)] \\
    \delta &= k^{*}\delta
    \end{aligned}
\end{equation}

Actually, the reason why catastrophic overfitting happens remains unclear, regardless of these previous works mentioned. And all previously proposed solutions result in further computational overhead compared to FGSM.

In this section, we introduce different adversarial training methods. Though PGD can lead to a robust model, it is computationally expensive. One gradient step methods, such as FGSM and RS-FGSM,  are computationally efficient but suffer from catastrophic overfitting which will cause robust accuracy against strong attacks fails to 0. Although there are some works trying to understand the catastrophic overfitting problem, the reason why it happens remains unclear. In this project, we will focus on understanding catastrophic overfitting in adversarial training. In the next section, we will introduce the catastrophic overfitting phenomenon in detail.

\section{Catastrophic overfitting in adversarial training}
In the last section, we introduced a family of defense methods called adversarial training, and show one of the most stable and robust methods in this family is PGD. However PGD has huge computational overhead and hard to be applied to large deep neural networks. But when we turn to computational efficient, one gradient step FGSM methods, we encounter the problem named catastrophic overfitting. In this section, we first define the formal definition of catastrophic overfitting and then demonstrate that catastrophic overfitting is a general phenomenon in adversarial training which is not limited to FGSM but also happens when we use $\mbox{DF}^{\infty}$-1 adversarial training.

\subsection{What is catastrophic overfitting}
\textbf{Catastrophic overfitting} During the adversarial training, we construct adversarial examples using an adversarial attack method, named method-A, such as FGSM. Then we evaluate the robust accuracy on the test dataset using another stronger adversarial attack method, named method-B, such as PGD. After a certain epoch, there is a sudden increase of method-A accuracy and a sudden decrease of method-B accuracy on both training and testing dataset. We call this phenomenon catastrophic overfitting. Catastrophic overfitting is not the same as overfitting to training dataset but is the overfitting to the weaker adversarial attack method-A. 
\begin{figure}[H]
 \begin{center}
  \includegraphics[width=0.8\textwidth]{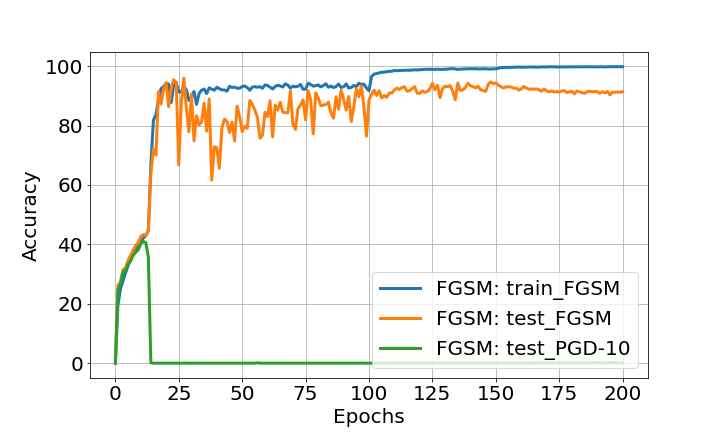}
  \caption{Visualization of the training process regarding FGSM adversarial training using PreAct-ResNet18 on CIFAR10 with $\epsilon$ = 8/255. Catastrophic overfitting happens around epoch 13 and is observed by a sudden increase of FGSM accuracy on the both training and testing dataset and a sudden decrease of PGD-10 accuracy on testing dataset.}
  \label{co_fgsm}
 \end{center}
\end{figure}

As shown on Figure \ref{co_fgsm}, the PreAct-ResNet18 is trained by FGSM on CIFAR10 with $\epsilon=8/255$. Catastrophic overfitting happens around epoch 13 and results in a sudden decrease of PGD-10 accuracy on the testing dataset and a sudden increase of FGSM accuracy on both training and testing datasets. In the next subsection, we will show that catastrophic overfitting is a general phenomenon for adversarial training and not only limited to vanilla FGSM.

\subsection{Catastrophic overfitting: a general phenomenon for adversarial training }
We first introduce the detailed experiment settings used in this project before showing more experiment results. Then we demonstrate that catastrophic overfitting still happens in RS-FGSM when $\epsilon$ or step size $\alpha$ is large. 

\textbf{Experiment settings} We train the PreAct-ResNet18 model from scratch on CIFAR10 with batch size=256, epoch=200 and initial learning rate=0.1. And learning rate will be decayed at epoch 100 and 150 by 10. The paper \cite{rice2020overfitting} find adversarial training will overfit to the training set and the best model is not the final model we obtain at the end of the training. In this project, we either evaluate the final model without early stopping to indicate whether catastrophic overfitting happens or not or we will select the best model based on test PGD-10 accuracy during the training process. We use PGD-50-10 (50 iterations and 10 restarts with step size $\alpha=\epsilon/4$) to evaluate the robustness of trained model.

\subsubsection{RS-FGSM suffers from catastrophic overfitting}
In paper \cite{wong2020fast}, the author trains the PreAct-ResNet18 model on CIFAR10 with $\epsilon=8/255$ using cyclic learning rate, and finds using step size $\alpha=1.25\epsilon$ leads to the best robust accuracy. And the model starts to suffer from catastrophic overfitting if further increasing the step size $\alpha$. We use the piecewise decay learning rate as suggested by paper\cite{rice2020overfitting} that can get the best robust accuracy among other learning rates. Here we run the experiment to check whether large step size $\alpha$ with fixed $\epsilon$ still suffers from catastrophic overfitting using piecewise decay learning rate and check whether this best ratio ($\alpha/\epsilon$) 1.25 will change according to different $\epsilon$ or not.

\begin{figure}[H]
\begin{center}
    \includegraphics[width=0.8\textwidth]{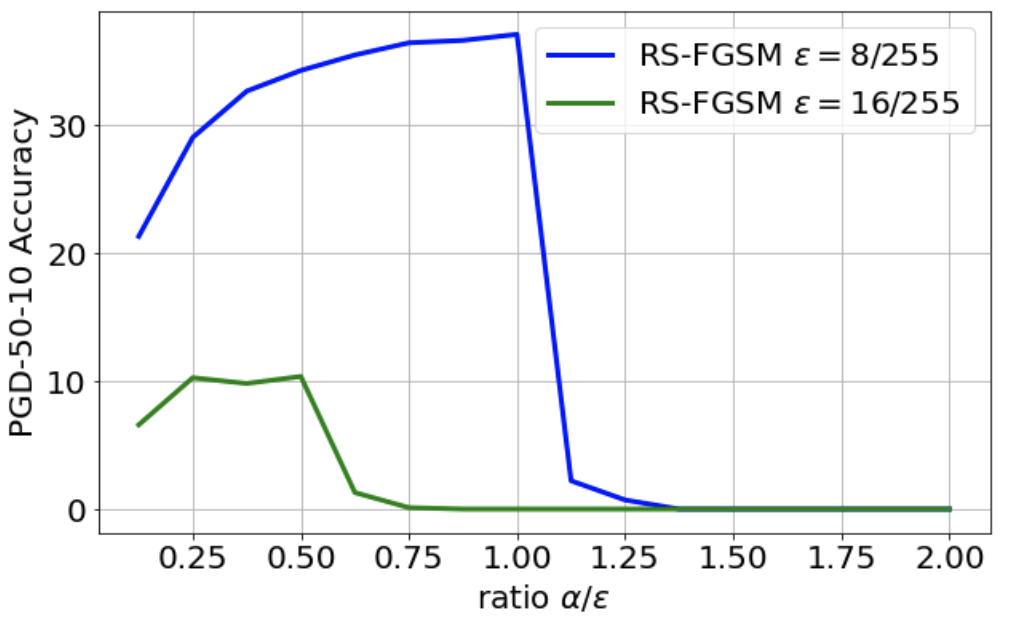}
    \caption{PGD-50-10 accuracy of RS-FGSM adversarial training over different step-size $\alpha=\mbox{ratio}\epsilon$ with fixed $\epsilon=8/255$ or $\epsilon=16/255$. We evaluate the PGD-50-10 accuracy of final model without early stopping.}
    \label{step-size}
\end{center}
\end{figure}

\textbf{Different step-size $\alpha$ and fixed $\epsilon$.} As shown on the Figure \ref{step-size}, we train the model using RS-FGSM adversarial training over different ratio($\alpha / \epsilon$) with fixed $\epsilon$ on CIFAR10. We observe that for $\epsilon=8/255$, catastrophic overfitting happens when $\alpha / \epsilon > 1$, while for $\epsilon=16/255$, catastrophic overfitting happens when $\alpha / \epsilon > 0.5$. Based on experiment results, we would say catastrophic overfitting happens regardless of the learning rate schedule and the best ratio will change according to different $\epsilon$. However, it's hard to tune this ratio each time for different $\epsilon$ when we use RS-FGSM. We will use $\alpha=\epsilon$ in this project.

\begin{figure}[H]
    \begin{center}
    \subfloat[Final model]{
      \includegraphics[width=0.5\textwidth]{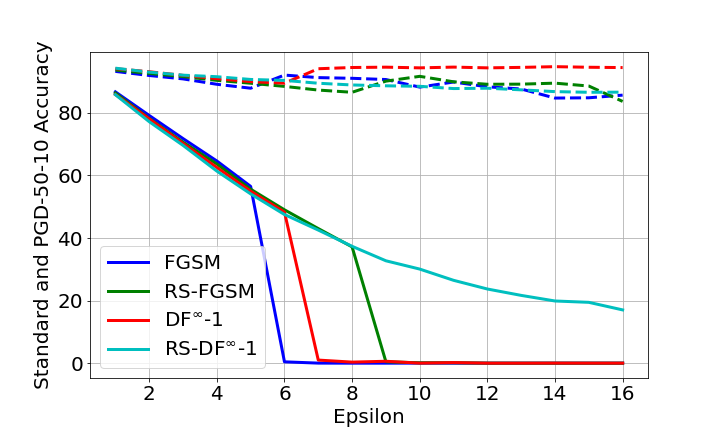}
    }
    \subfloat[Best model]{
      \includegraphics[width=0.5\textwidth]{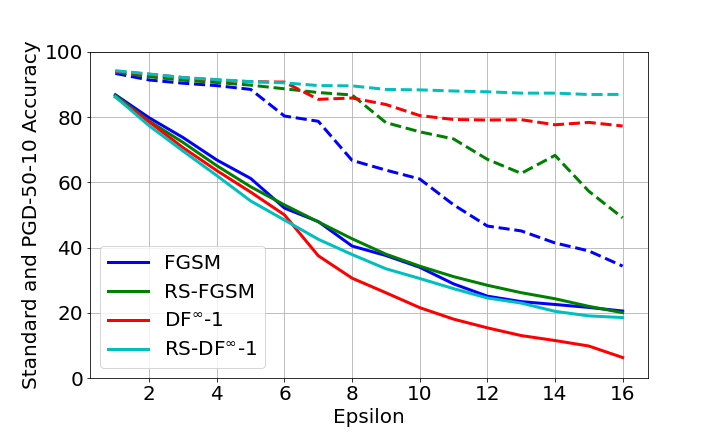}
    }
    \caption{Standard accuracy (dashed line) and PGD-50-10 accuracy (solid line) from different adversarial training (AT) methods with different $\epsilon$. For RS-FGSM, the step-size $\alpha=\epsilon$. For $\mbox{DF}^{\infty}$-1, the overshoot $\eta$ is fixed to 0.02}
    \label{epsilon}
    \end{center}
\end{figure}

In paper \cite{andriushchenko2020understanding}, the author finds RS-FGSM still suffer from catastrophic overfitting on CIFAR10 when $\epsilon > 9/255$  with fixed $\alpha=1.25\epsilon$ and the model is trained by cyclic learning rate. Here we run the experiments over \textbf{different $\epsilon$ with fixed step size $\alpha=\epsilon$} using piecewise decay learning rate. As shown on Figure \ref{epsilon}, RS-FGSM starts to suffer from CO when $\epsilon > 8/255$. This is larger compared to vanilla FGSM which starts to suffer from catastrophic overfitting when $\epsilon > 5/255$. So we would say RS-FGSM does not prevent catastrophic overfitting for large $\epsilon$, but it does mitigate the catastrophic overfitting problem by extending to larger $\epsilon$ compared to vanilla FGSM.
    
Based on these two experiments, we could conclude that RS-FGSM still suffers from catastrophic overfitting, but extends the working range of $\epsilon$ compared to vanilla FGSM. And if we want to avoid catastrophic overfitting, we can decrease the step size $\alpha$, but this will leads to a sub-optimal solution, the perturbations we find using a smaller step size $\alpha$ will be in a smaller $l_\infty$-ball than the one used during the evaluation.

\subsubsection{$\mbox{DF}^{\infty}$-1 suffers from catastrophic overfitting}
In addition to FGSM and RS-FGSM, we also train the model using 1-iteration $l_\infty$ DeepFool method ($\mbox{DF}^{\infty}$-1, DF's superscript of $\infty$ means using $l_\infty$ norm to calculate perturbation in each iteration)\cite{moosavidezfooli2016deepfool}. The reason we use 1 iteration DeepFool instead of more iterations is because we want to make this method have comparable computational efficiency as one gradient step FGSM. The difference between FGSM and $\mbox{DF}^{\infty}$-1 is that FGSM always has the fixed step size $\alpha$ for all input examples while $\mbox{DF}^{\infty}$-1 will adapt the length of perturbation dynamically for each input example and the perturbation will not overly perturb to decision boundary, as shown on Equation \ref{DF-1}. 

\begin{equation}\label{DF-1}
     \delta_{DF^{\infty}-1} = (1+\eta)DF(x)
\end{equation}

\begin{equation}\label{RS-DF-1}
\begin{aligned}
    \delta &\sim \mathcal{U}([-\epsilon, \epsilon]^d) \\\
    \delta_{RS-DF^{\infty}-1} &= \textstyle\prod_{[-\epsilon, \epsilon]^d}(\delta +(1+\eta)DF(x+\delta))
\end{aligned}
\end{equation}

We can also add random initialization to $\mbox{DF}^{\infty}$-1 as shown on Equation \ref{RS-DF-1}.

\begin{figure}[H]
 \begin{center}
  \includegraphics[width=0.8\textwidth]{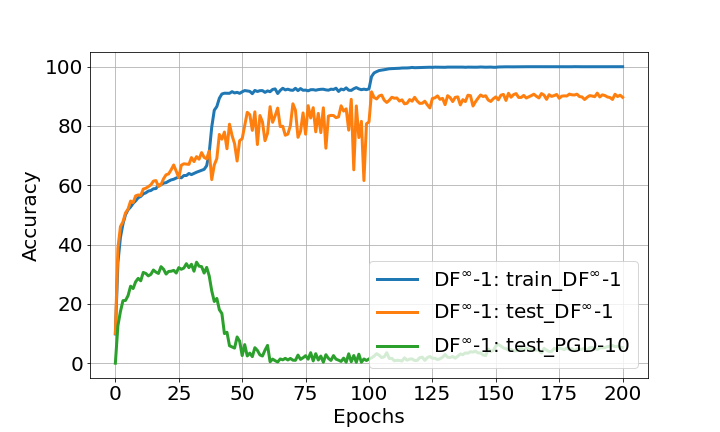}
  \caption{Visualization of the training process regarding $\mbox{DF}^{\infty}$-1 adversarial training using PreAct-ResNet18 on CIFAR10 with $\eta$ = 0.02. Catastrophic overfitting happens around epoch 38.}
  \label{co_df}
 \end{center}
\end{figure}

As shown on Figure \ref{co_df}, the PreAct-ResNet18 model trained by $\mbox{DF}^{\infty}$-1 on CIFAR10 with $\epsilon=8/255$ suffers from catastrophic overfitting. There is a decrease of PGD10 accuracy from robust to 0 on testing dataset and an increase of $\mbox{DF}^{\infty}$-1 accuracy on both training and testing dataset in few epochs. And we observe that the speed of decrease and increase is slower than FGSM. We will discuss the difference between $\mbox{DF}^{\infty}$-1 and FGSM in next section. As shown on Figure \ref{epsilon}, when we decrease the $\epsilon$ to 6/255, $\mbox{DF}^{\infty}$-1 can train a robust model without suffering from catastrophic overfitting. When we add random initialization to $\mbox{DF}^{\infty}$-1, we can extend the $\epsilon$ to 16/255 at which there is no catastrophic overfitting happens but leads to sub-optimal robust accuracy. 

In this section, we define the formal definition of catastrophic overfitting and empathize that this is overfitting between weak attacks and strong attacks intead of overfitting between training and testing dataset. Besides, we demonstrate that catastrophic overfitting is a general phenomenon in adversarial training and suffers not only by FGSM, RS-FGSM, but also by $\mbox{DF}^{\infty}$-1. In next section we will analysis catastrophic overfitting in geometric way and discuss the difference between FGSM and $\mbox{DF}^{\infty}$-1.

\section{Geometric analysis of catastrophic overfitting}
In the last section, we find catastrophic overfitting happens not only in FGSM and but also in $\mbox{DF}^{\infty}$-1 adversarial training methods. And once it happens, robustness of the model will drop to 0 in a few epochs. Thus it is important to analyze what happens before and after catastrophic overfitting. In this section, we first analyze the decision boundaries of the classifiers trained by FGSM or $\mbox{DF}^{\infty}$-1 before and after catastrophic overfitting separately. Then we compare the difference between them. We find although these two methods both cause catastrophic overfitting, they lead to totally different geometric properties after catastrophic overfitting. 

\subsection{Geometric analysis of FGSM adversarial training}

We first introduce \textbf{DeepFool} \cite{moosavidezfooli2016deepfool} before going deeper into the geometric analysis of FGSM adversarial training. 
\begin{equation}\label{deepfool}
    \Delta(x;\hat{k})=\displaystyle\min_{r}\|r\|_2 \mbox{ subject to } \hat{k}(x+r)\neq \hat{k}(x)
\end{equation}
$\mbox{DF}^2$ (DF's superscript of 2 means using $l_2$ norm to calculate perturbations in each iteration) is an algorithm used to find the smallest perturbations which can fool the deep networks based on iterative linearization of the classifier, as shown in equation \ref{deepfool}.
If the decision boundary is linear, $\mbox{DF}^2$ only needs 1 iteration to find the smallest perturbation, which is perpendicular to the decision boundary. The more complex and non-linear decision boundary is, the more iterations $\mbox{DF}^2$ are needed to find the valid perturbations. In this way, we can use the number of $\mbox{DF}^2$ iterations to estimate the complexity of the decision boundary and use $l_2$ norm of the $\mbox{DF}^2$ perturbations to estimate the robustness of the classifier. During the implementation, we set a maximum of 50 iterations to avoid $\infty$ loop. Based on the empirical results from the paper \cite{moosavidezfooli2016deepfool}, $\mbox{DF}^2$ converges in a few iterations (i.e., less than 3), thus 50 iterations should be enough to find the valid perturbation for almost all data points.

The differences between $\mbox{DF}^2$ we use to evaluate the robustness of classifiers here and $\mbox{DF}^{\infty}$-1 we use in the previous section to train the robust model can be summarized into the two following aspects: (1) the number of the iterations is different. $\mbox{DF}^2$ either stops looping when finding the valid perturbation that can deceive the classifier or reaches the maximum of 50 iterations, while $\mbox{DF}^{\infty}$-1 always stops after 1 iteration no matter whether finding the valid perturbation or not; (2) $\mbox{DF}^2$ uses $l_2$ norm to calculate the perturbation in each iteration, while $\mbox{DF}^{\infty}$-1 uses $l_\infty$ norm to be coincident with the $l_\infty$ threat model we use in this project.

Now, we will introduce the way we visualize the decision boundary. Since it is impossible to visualize the decision boundary of a high-dimensional classifier, we draw the cross-section of the decision boundary spanned by two vectors. One vector is the perturbation vector calculated by $\mbox{DF}^2$, and the other is the perturbation vector calculated by the adversarial method used in the training process (FGSM or $\mbox{DF}^{\infty}$-1). We use green color to represent the true class and different red color to represent false classes. The origin represents the clean inputs. 

\begin{figure}[H]
\begin{center}
\subfloat[Epoch 10]{
  \includegraphics[width=0.4\textwidth]{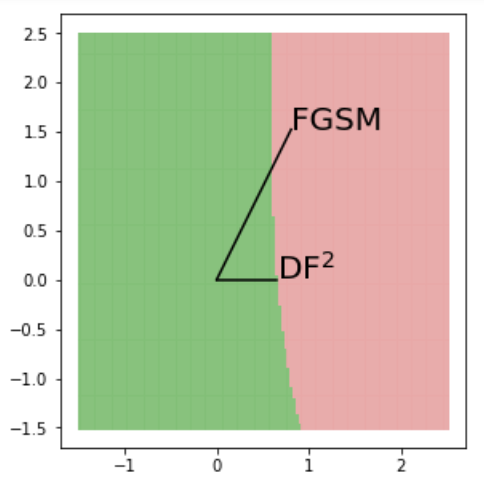}
}
\subfloat[Epoch 15]{
  \includegraphics[width=0.4\textwidth]{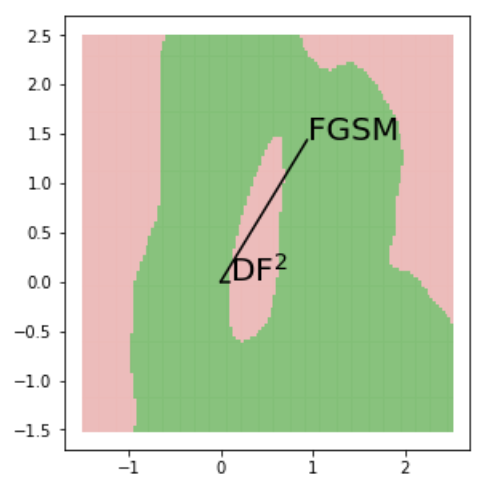}
}
  \caption{Show the cross-section of the decision boundary spanned by two vectors. One is calculated by $\mbox{DF}^2$, and the other one is calculated by the FGSM adversarial method used in the training process. Green color represents the true class and the different red color represents false classes. The original point represents the clean inputs. The classifier is trained by FGSM with $\epsilon=8/255$ and catastrophic overfitting happens at epoch13. Here epoch10 and epoch15 are the snapshots of the classifier before and after catastrophic overfitting separately.}
\label{fgsm_decision_boundary}
\end{center}
\end{figure}

As shown in Figure \ref{co_fgsm}, the PreAct-ResNet18 model trained by vanilla FGSM with $\epsilon=8/255$ has catastrophic overfitting occurring at epoch 13. We draw the cross-sections of the decision boundary of a specific input example at epoch10 and epoch15 separately. As shown in Figure \ref{fgsm_decision_boundary}, we observe that (1) small perturbation becomes more effective than the large one along the FGSM direction after catastrophic overfitting. At epoch 10, the small perturbation cannot find the effective adversarial example while the large perturbation can. At epoch15, small perturbation is more effective than large perturbation along the FGSM direction. (2) clean input becomes much close to the decision boundary after catastrophic overfitting. At epoch10, the smallest perturbation found by $\mbox{DF}^2$ that can fool the network is much larger than the one found at epoch15.

\begin{figure}[H]
\begin{center}
\subfloat[]{
  \includegraphics[width=0.49\textwidth]{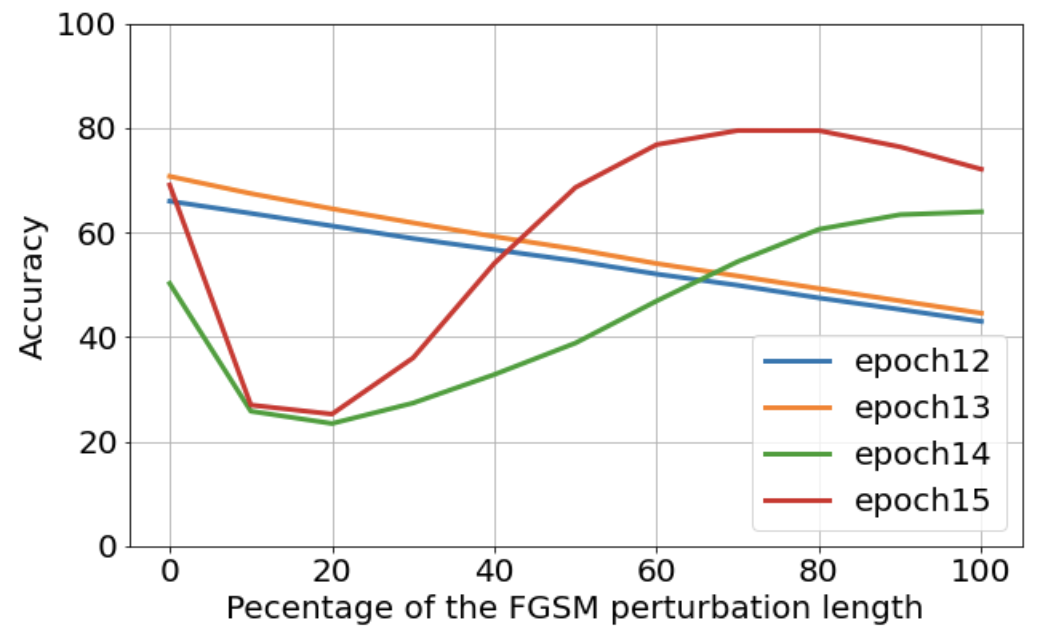} \label{fgsm_acc_of_diff_len}
}
\subfloat[]{
  \includegraphics[width=0.49\textwidth]{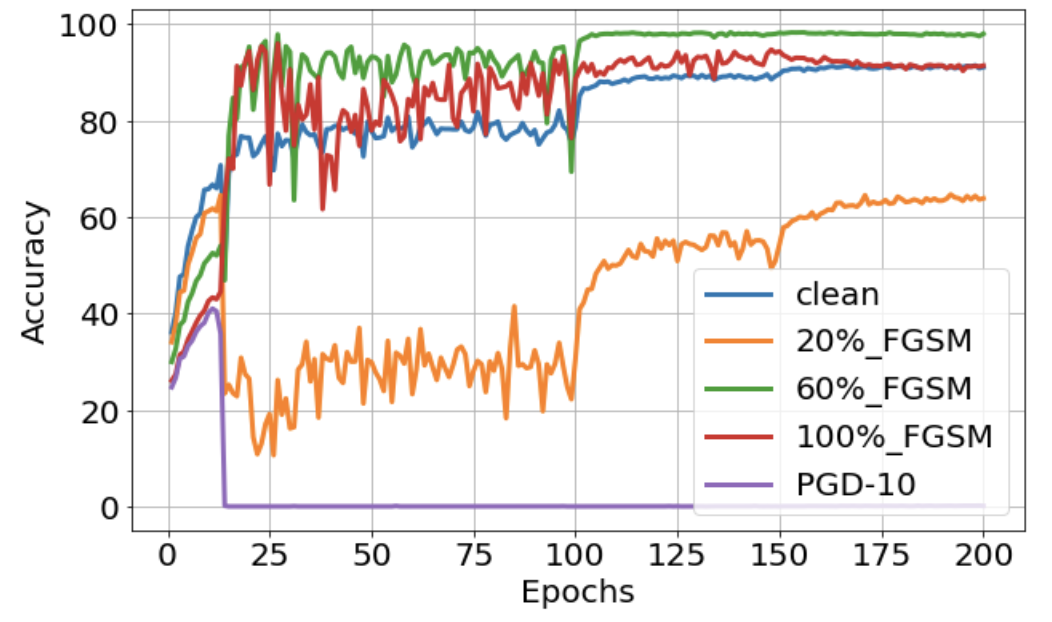} \label{fgsm_acc_of_diff_len_along_epoch}
}
\caption{The model is PreAct-ResNet18 trained by vanilla FGSM with $\epsilon=8/255$ and occurs catastrophic overfitting at epoch13 as shown on Figure \ref{co_fgsm} (a) The test accuracy of perturbed points generated by different length of FGSM perturbations.(b) The test accuracy of perturbed points generated by different length of FGSM perturbations or PGD-10 perturbations during the training process.}
\label{fgsm_geometric_curve}
\end{center}
\end{figure}  

Apart from qualitative analysis, we also add quantitative analysis to support our observations. In order to verify that small perturbation is more effective than large perturbation after catastrophic overfitting, we do two experiments as shown in Figure \ref{fgsm_geometric_curve}. For a fixed epoch, we calculate the test accuracy of perturbed inputs generated by different lengths of FGSM perturbation. The result shown in Figure \ref{fgsm_acc_of_diff_len} is coincident with the observation we have by drawing the cross-section of the decision boundary. The curves of epoch12 and epoch13 show that before CO, large perturbation has smaller accuracy than small perturbation and thus it is more effective. And the curves of epoch14 and epoch15 show that after CO, we have the worst accuracy at about 20\% of the FGSM perturbation length, and then the accuracy increases as the length of the perturbation increases, which means large perturbation becomes less effective than small perturbation. In order to make sure that small perturbation is more effective than large perturbation holds for all epochs after catastrophic overfitting, we calculate the accuracy of perturbed points generated by 20\%, 60\%, or 100\% FGSM perturbation length. As shown in Figure \ref{fgsm_acc_of_diff_len_along_epoch}, after CO, the perturbed points generated by 20\% FGSM perturbation length always have the smallest accuracy which means these perturbations are the most effective. For a robust model, small perturbation is weaker than large perturbation, but after catastrophic overfitting, we have the counter-intuitive observations which reflect that new decision boundary may be generated around the perturbed points to make them be classified successfully.

In order to verify that the clean input becomes much near to the decision boundary after CO, we calculate the expected $l_2$ norm of $\mbox{DF}^2$ perturbations. As shown in Figure \ref{df_norm}, we observe that after CO happens, the expected $l_2$ norm of $\mbox{DF}^2$ perturbations decreases suddenly.

\begin{figure}[H]
 \begin{center}
  \includegraphics[width=0.7\textwidth]{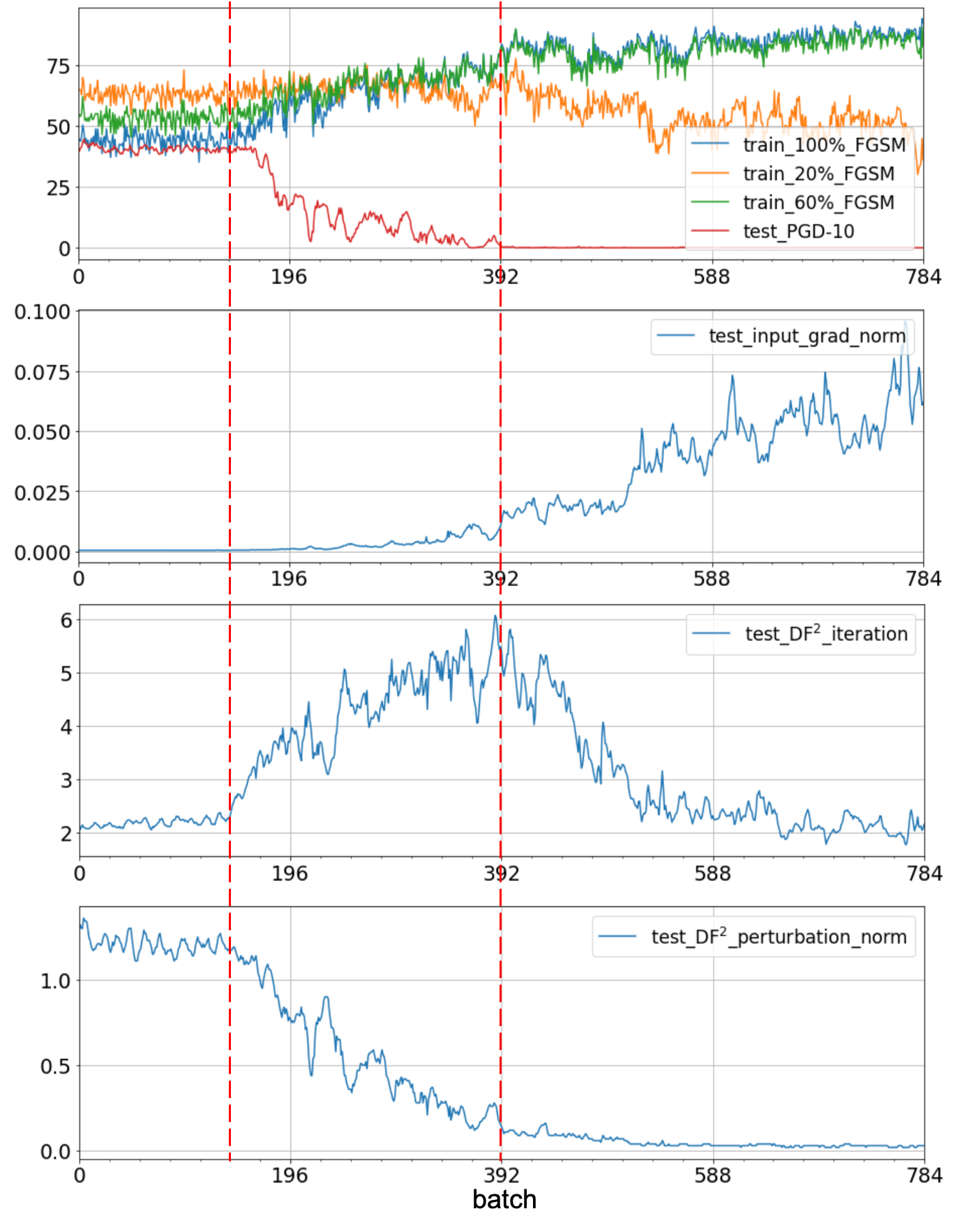}
  \caption{We train PreAct-ResNet18 using FGSM on CIFAR10 with $\epsilon=8/255$ and on figure \ref{co_fgsm} we show CO happens at epoch13. Thus, we resume the training from epoch12 and train for another 4 epochs and then evaluate the model by batch. Each epoch has 196 batches. }
  \label{fgsm_by_batch}
 \end{center}
\end{figure}

Based on the previous analysis, we clearly demonstrate the geometric difference between classifiers before and after catastrophic overfitting. For FGSM adversarial training, the catastrophic overfitting happens almost within one epoch. We want to check what happens in these particular epochs around catastrophic overfitting. We train PreAct-ResNet18 using vanilla FGSM on CIFAR10 with $\epsilon=8/255$ and on Figure \ref{co_fgsm} we show CO happens at epoch13. Thus, we resume the model from epoch12 and train for another 4 epochs (the number of batches in one epoch is 196). Then, we evaluate the accuracy, $l_2$ norm of the input gradients, the number of $\mbox{DF}^2$ iterations, and $l_2$ norm of $\mbox{DF}^2$ perturbations by batch. The result is shown in Figure\ref{fgsm_by_batch}. We observe that (1) From batch 0 to 140, the model is robust based on the test PGD10 accuracy. At that time, small FGSM perturbations are less effective than large FGSM perturbations. (2) From batch 140 to 392, the number of $\mbox{DF}^2$ iterations increases quickly and the $l_2$ norm of the $\mbox{DF}^2$ perturbations decreases correspondingly. During this stage, the model starts to lose robustness and the test PGD10 accuracy drops to zero. And we also notice that the accuracy of perturbed points constructed by different lengths of FGSM perturbation all increase and have almost the same accuracy. This gives us a clue that FGSM direction becomes ineffective at this stage.(3) From batch 392 to 784, the test PGD10 accuracy remains at 0; $l_2$ norm of the input gradients increases quickly; the accuracy of perturbed points generated by 20\% of FGSM perturbation starts to decrease, which means small perturbations become more effective than large perturbations; new decision boundary is generated around the perturbed points.

\begin{figure}[H]
\begin{center}
\subfloat[Batch 60]{
  \includegraphics[width=0.24\textwidth]{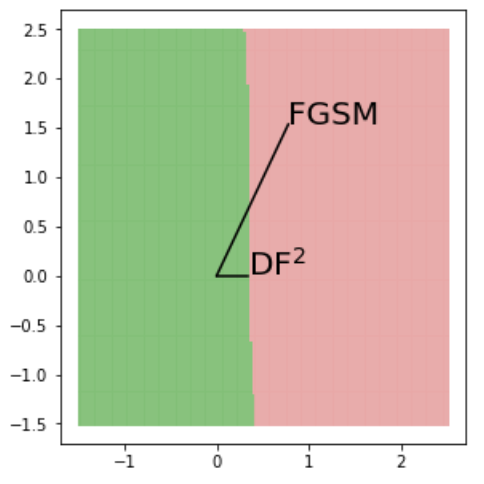} \label{batch60}
}
\subfloat[Batch 240]{
  \includegraphics[width=0.24\textwidth]{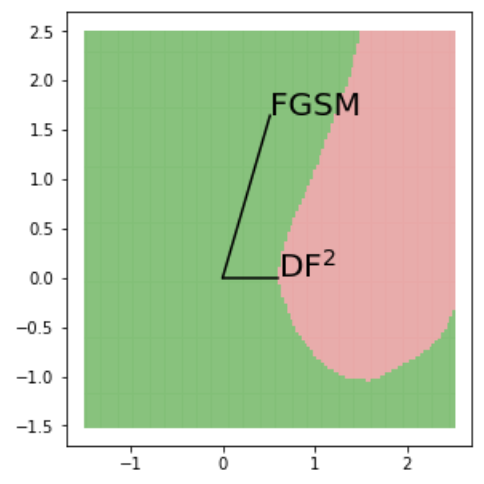} \label{batch240}
}
\subfloat[Batch 360]{
  \includegraphics[width=0.24\textwidth]{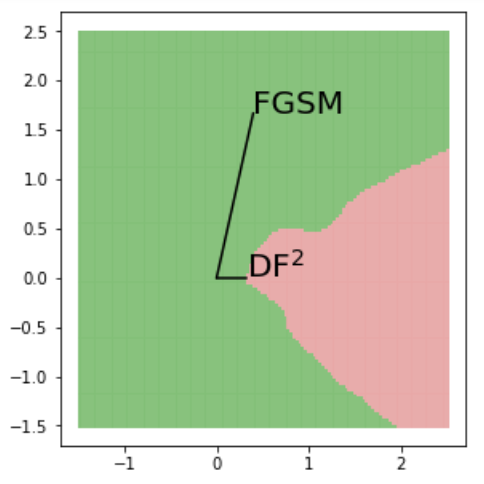} \label{batch360}
}
\subfloat[Batch 780]{
  \includegraphics[width=0.24\textwidth]{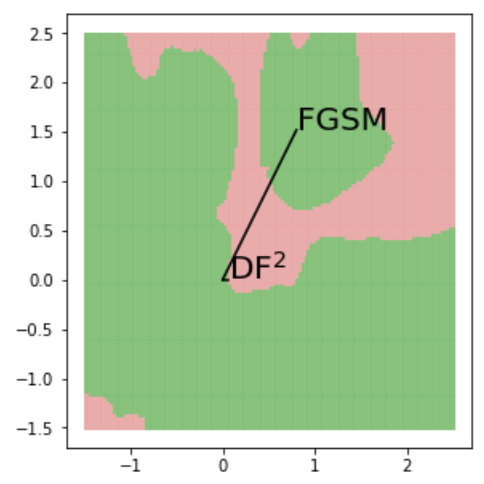} \label{batch780}
}
\caption{Show the cross-section of the decision boundary spanned by perturbation vectors, which are calculated by $\mbox{DF}^2$ and FGSM adversarial method used in the training process at different batches.}
\label{decision boundary by batch}
\end{center}
\end{figure}

We also draw the cross-sections of the decision boundaries spanned by perturbation vectors, which are calculated by $\mbox{DF}^2$ and FGSM adversarial method used in the training process at different batches as shown on Figure \ref{decision boundary by batch}. The visualization of the decision boundary corresponds to our analysis. From the Figure \ref{batch240}, \ref{batch360}, we observe that at batch 240 and batch 360, all perturbations along the FGSM direction is ineffective. And at batch 60 before CO, large perturbation is more effective, while at batch 780 after CO, small perturbation is more effective.

\subsection{Geometric analysis of $\mbox{DF}^{\infty}$-1 adversarial training}
\begin{figure}[H]
\begin{center}
\subfloat[Epoch 36]{
  \includegraphics[width=0.36\textwidth]{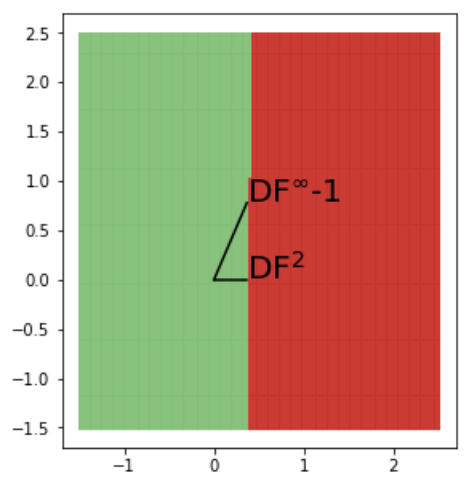}
}
\subfloat[Epoch 70]{
  \includegraphics[width=0.36\textwidth]{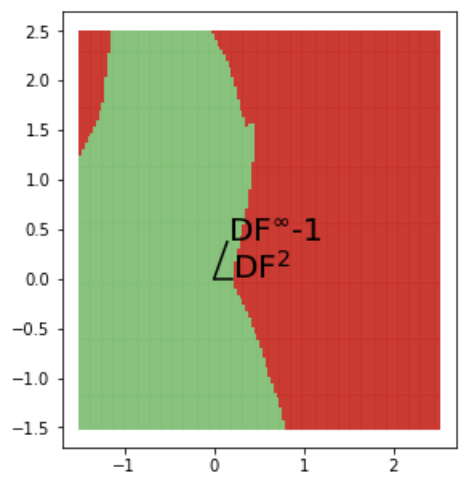}
}
\caption{ Show the cross-section of the decision boundary spanned by two vectors. One is calculated by $\mbox{DF}^2$ and the other one is calculated by $\mbox{DF}^{\infty}$-1 used in the training process. The classifier is trained by $\mbox{DF}^{\infty}$-1 with $\epsilon=8/255$. Epoch36 and epoch70 are the snapshots of the classifier before and after CO separately.}
\label{DF decision boundary}
\end{center}
\end{figure}

\begin{figure}[H]
 \begin{center}
  \includegraphics[width=0.8\textwidth]{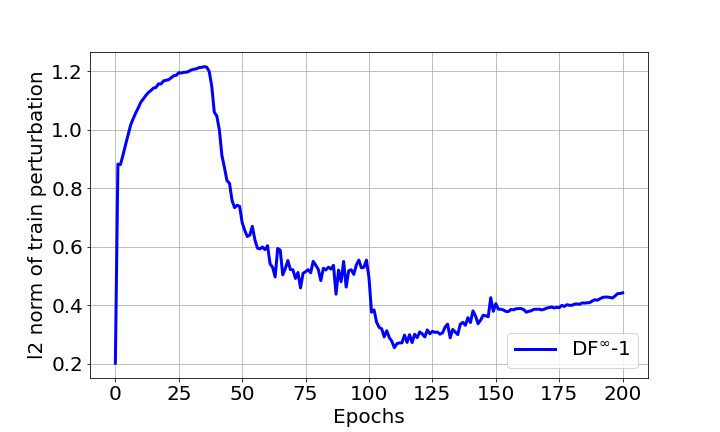}
  \caption{The expected $l_2$ norm of perturbations calculated by $\mbox{DF}^{\infty}$-1 during the training process}
  \label{train_perturbation_norm}
 \end{center}
\end{figure}

\begin{figure}[H]
\begin{center}
\subfloat[]{
  \includegraphics[width=0.5\textwidth]{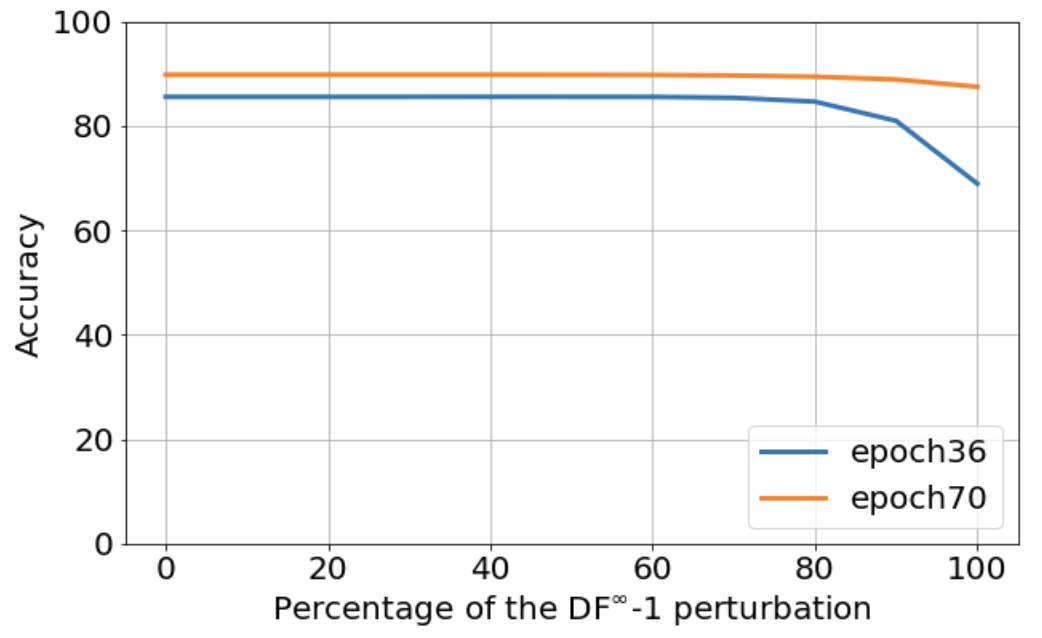} \label{df_acc_of_diff_len}
}
\subfloat[]{
  \includegraphics[width=0.5\textwidth]{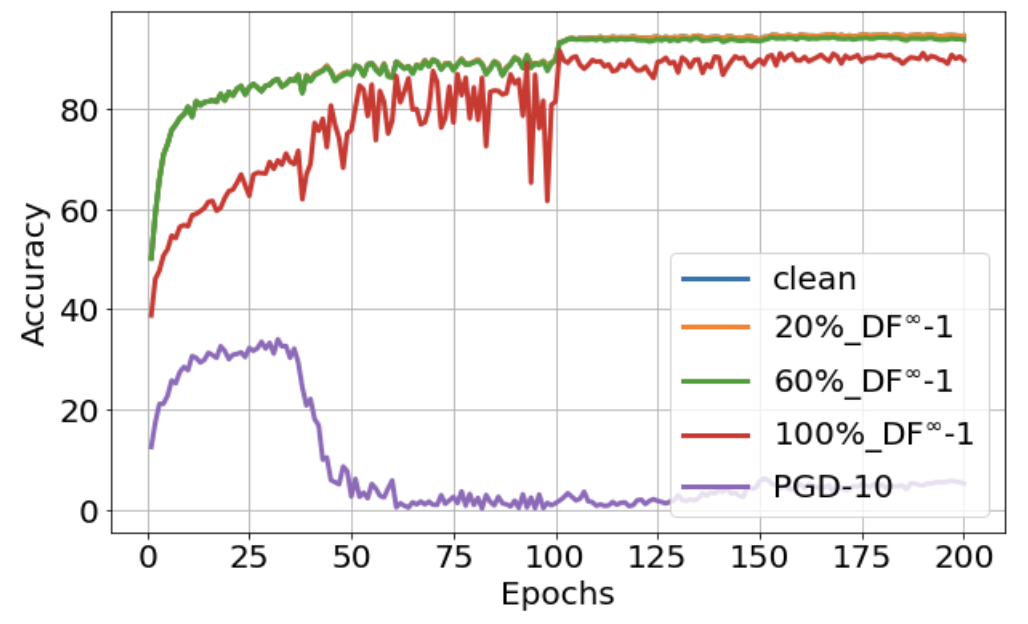} \label{df_acc_of_diff_len_along_epoch_df_df}
}
\caption{The model is PreAct-ResNet18 trained by $\mbox{DF}^{\infty}$-1 with $\epsilon= 8/255$ and starts to suffer from CO at epoch 36 as shown in Figure \ref{co_df}. (a) The test accuracy of perturbed points generated by different length of $\mbox{DF}^{\infty}$-1 perturbations. (b) The test accuracy of perturbed points generated by different length of $\mbox{DF}^{\infty}$-1 perturbations or PGD-10 perturbations during the training process.}
\label{DF geometric curve}
\end{center}
\end{figure}

From the previous section we know catastrophic overfitting is not only limited to FGSM but also suffered by $\mbox{DF}^{\infty}$-1. Here we apply the same geometric analysis for $\mbox{DF}^{\infty}$-1 as we do for FGSM in the last subsection. As shown in Figure \ref{co_df}, $\mbox{DF}^{\infty}$-1 with $\epsilon=8/255$ starts to occur CO at epoch 36, which means the test PGD-10 accuracy starts to decrease. And at around epoch 65, the accuracy decreases to 0. As shown on Figure \ref{DF decision boundary}, we draw the cross-sections of the decision boundary of a specific input example at epoch 36 and epoch 70 separately. And we also add the quantitative analysis on Figure \ref{train_perturbation_norm}, \ref{DF geometric curve} to support our analysis.

The expected $l_2$ norm of the perturbation calculated by $\mbox{DF}^{\infty}$-1 during the training process becomes smaller after CO, as shown in Figure \ref{train_perturbation_norm}. This is different from FGSM, whose perturbation length will not change during the training process once the step size is fixed.

As shown in Figure \ref{df_acc_of_diff_len}, \ref{df_acc_of_diff_len_along_epoch_df_df}, large percentage of $\mbox{DF}^{\infty}$-1 perturbation length is always more effective than small percentage of $\mbox{DF}^{\infty}$-1 perturbation length no matter CO happens or not. But after CO happens, perturbations generated by $\mbox{DF}^{\infty}$-1 become smaller, thus still lose their effectiveness. Besides, the clean input becomes much near to the decision boundary after CO. As shown in Figure \ref{DF decision boundary}, the smallest perturbation found by $\mbox{DF}^2$ which can fool the network at epoch 36 is larger than the one found at epoch 70. This can also be reflected on Figure \ref{df_norm} that after CO happens, the expected $l_2$ norm of $\mbox{DF}^2$ perturbations decreases.

\subsection{The difference between FGSM and $\mbox{DF}^{\infty}$-1}

\begin{figure}[H]
\begin{center}
\subfloat[accuracy]{
  \includegraphics[width=0.48\textwidth]{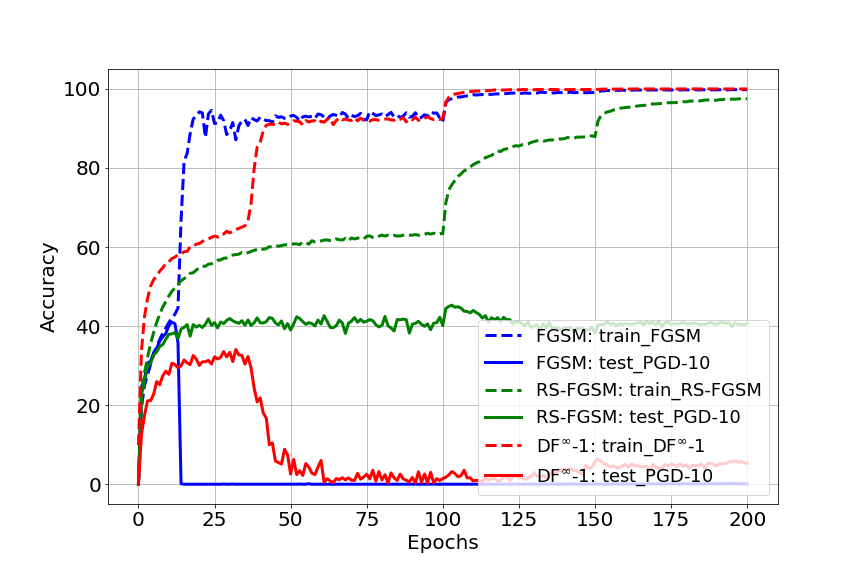} \label{acc_compare}
}
\subfloat[$l_2$ norm of input gradients]{
  \includegraphics[width=0.48\textwidth]{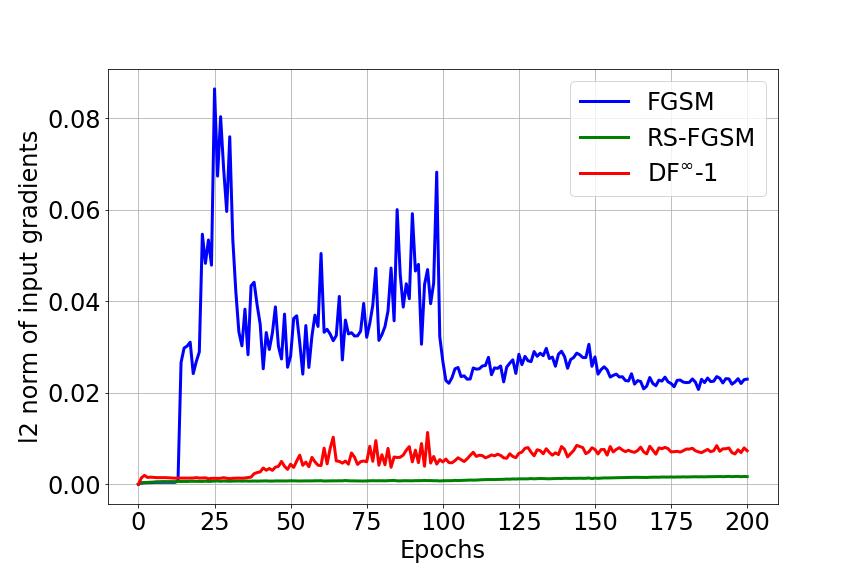} \label{input_gradient}
}
\newline
\subfloat[number of $\mbox{DF}^2$ iterations]{
  \includegraphics[width=0.48\textwidth]{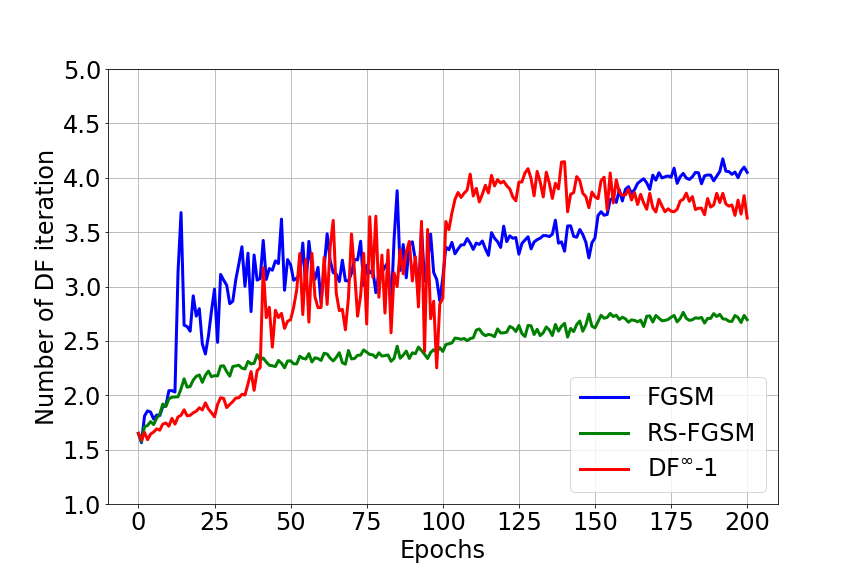} \label{df_loop}
}
\subfloat[$l_2$ norm of $\mbox{DF}^2$ perturbations]{
  \includegraphics[width=0.48\textwidth]{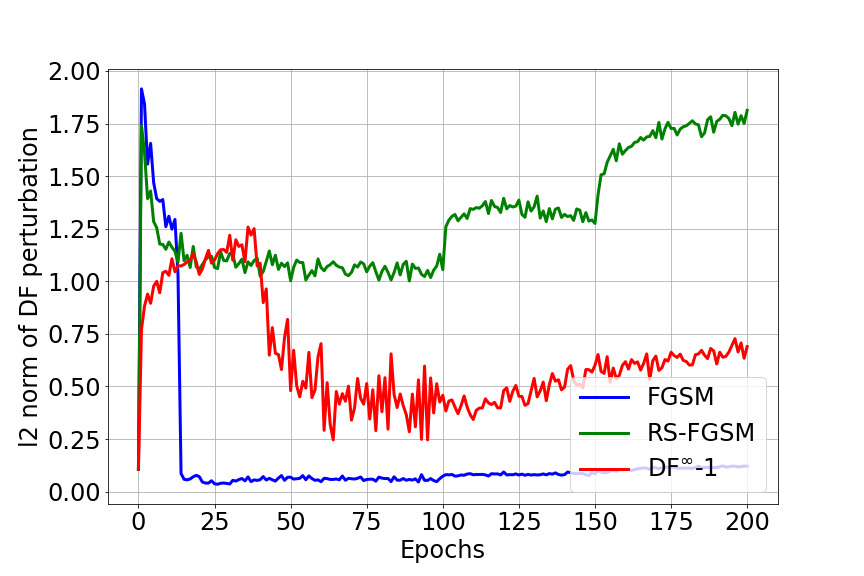} \label{df_norm}
}
\caption{Different observations during the training process, $\epsilon=8/255$ }
\label{observations}
\end{center}
\end{figure}

Based on previous analysis, we observe that although both FGSM and $\mbox{DF}^{\infty}$-1 adversarial training suffer from catastrophic overfitting, they have different geometric properties after catastrophic overfitting. We further compare FGSM, $\mbox{DF}^{\infty}$-1, RS-FGSM (not suffer from CO) under different perspectives. As shown in Figure \ref{acc_compare}, once CO happens, there is a sudden drop in PGD10 accuracy on testing dataset and a sudden increase in FGSM or $\mbox{DF}^{\infty}$-1 accuracy on training dataset. But PGD10 accuracy decreases much faster in FGSM compared to $\mbox{DF}^{\infty}$-1. This might because FGSM perturbation length will not decrease since the step size is fixed, while $\mbox{DF}^{\infty}$-1 perturbation length will decrease when CO happens as shown in Figure \ref{train_perturbation_norm},  and decreasing perturbation length can slow down the process of CO.

As shown on Figure \ref{input_gradient}, there is a sudden increase of the expected $l_2$ norm of input gradients for both FGSM and $\mbox{DF}^{\infty}$-1. However, FGSM is almost an order of magnitude larger than $\mbox{DF}^{\infty}$-1. As shown on Figure \ref{df_loop} \ref{df_norm}, the number of $\mbox{DF}^2$ iterations needed to calculate the smallest perturbations fooling the network also increases while the expected $l_2$ norm of $\mbox{DF}^2$ perturbations decreases. And the $l_2$ norm of $\mbox{DF}^2$ perturbations of the model trained by FGSM is much smaller than the model trained by $\mbox{DF}^{\infty}$-1 after CO. This might be because FGSM generates a new decision boundary near the clean inputs while $\mbox{DF}^{\infty}$-1 do not. 

\begin{figure}[H]
\begin{center}
\subfloat[]{\label{fgsm acc along fgsm}
  \includegraphics[width=0.53\textwidth]{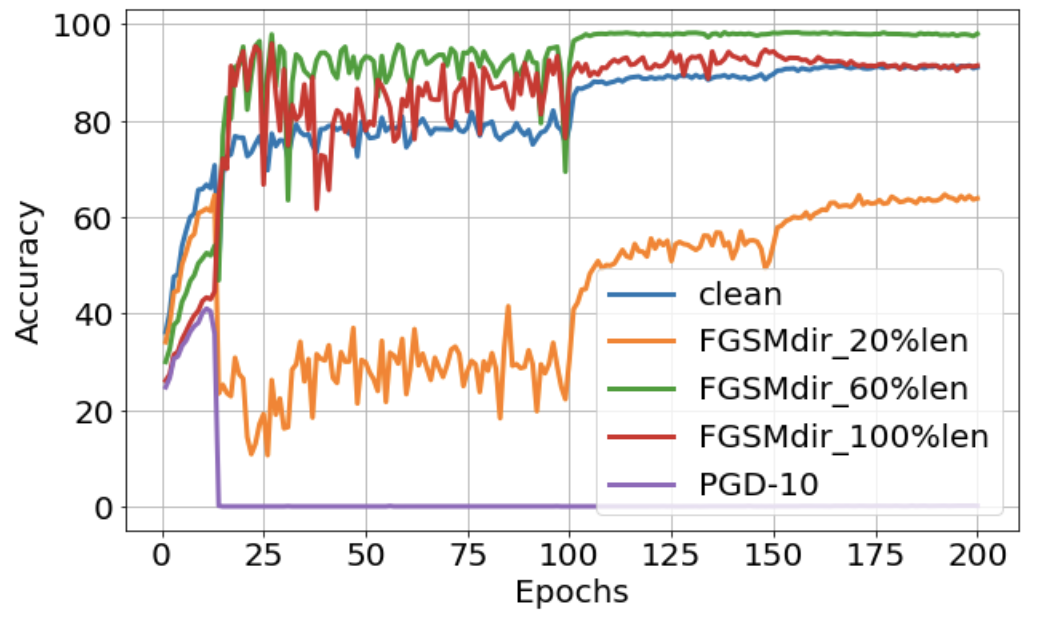}
}
\subfloat[]{\label{fgsm boundary along fgsm}
  \includegraphics[width=0.32\textwidth]{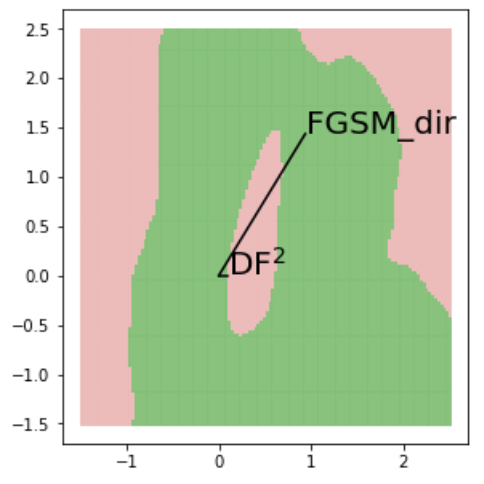}
}
\\
\subfloat[]{\label{fgsm acc along df}
  \includegraphics[width=0.53\textwidth]{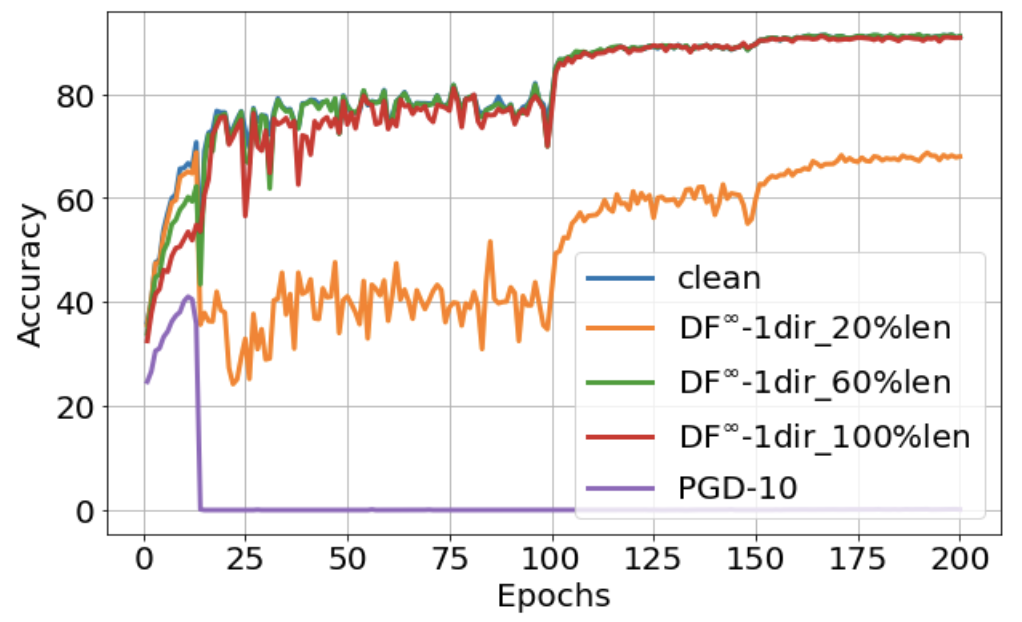}
}
\subfloat[]{\label{fgsm boundary along df}
  \includegraphics[width=0.32\textwidth]{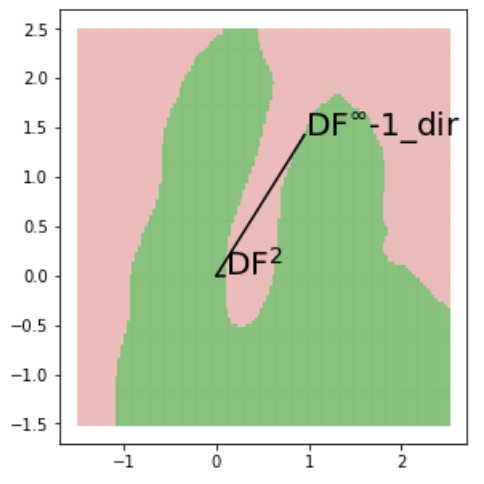}
}
\caption{Cross-sections of the decision boundaries and robust accuracy under different percentage of FGSM perturbation length along FGSM or $\mbox{DF}^{\infty}$-1 direction on the model trained by FGSM. (a) were tested along FGSM direction while (c) were tested along $\mbox{DF}^2$direction.}
\label{compare accuracy between fgsm}
\end{center}
\end{figure}
\begin{figure}[H]
\begin{center}
\subfloat[]{\label{df acc along fgsm}
  \includegraphics[width=0.53\textwidth]{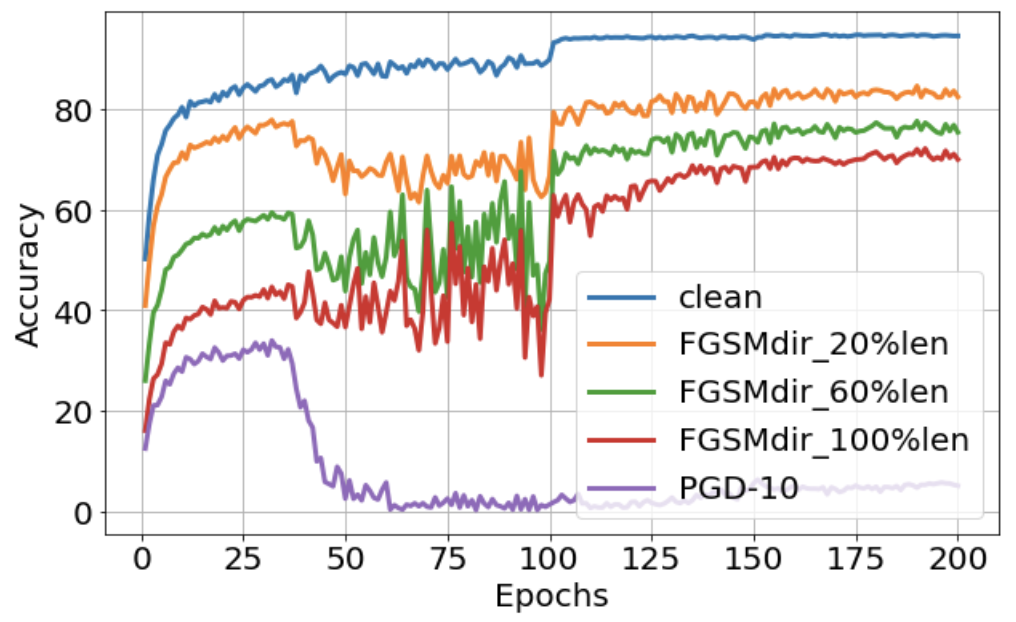}
}
\subfloat[]{\label{df boundary along fgsm}
  \includegraphics[width=0.32\textwidth]{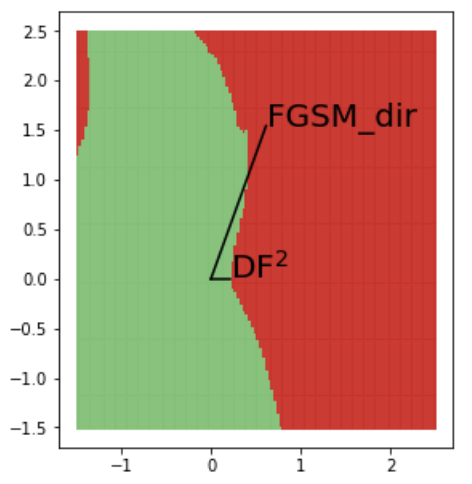}
}
\\
\subfloat[]{\label{df acc along df}
  \includegraphics[width=0.53\textwidth]{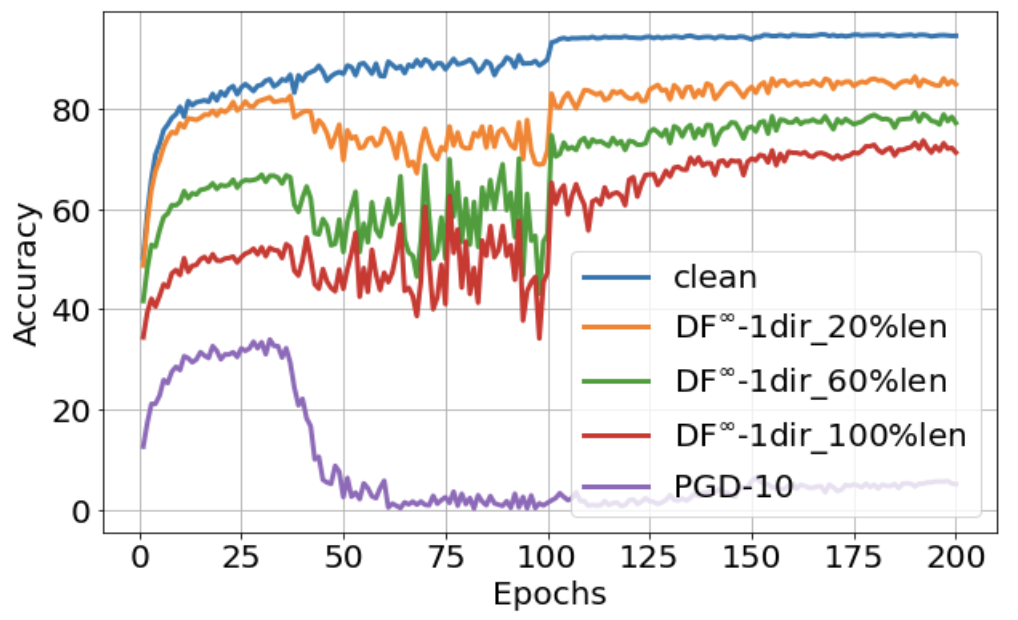}
}
\subfloat[]{\label{df boundary along df}
  \includegraphics[width=0.32\textwidth]{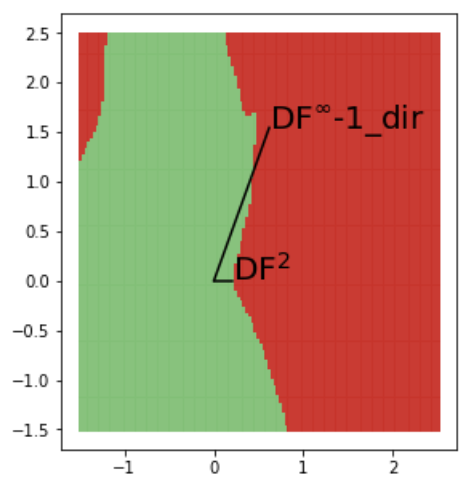}
}
\caption{Cross-sections of the decision boundaries and robust accuracy under different percentage of FGSM perturbation length along FGSM or $\mbox{DF}^{\infty}$-1 direction on model trained by $\mbox{DF}^{\infty}$-1. (a) were tested along FGSM direction while (c) were tested along $\mbox{DF}^2$direction.}
\label{compare accuracy between DF-1}
\end{center}
\end{figure}

We draw cross-sections of the decision boundaries and calculate robust accuracy under different lengths of perturbations along FGSM or $\mbox{DF}^{\infty}$-1 direction of the model trained by FGSM or $\mbox{DF}^{\infty}$-1. As shown on Figure \ref{compare accuracy between fgsm}, for \textbf{FGSM}, after CO happens, the small perturbation along FGSM and $\mbox{DF}^{\infty}$-1 direction becomes more effective than the large one, and new decision boundary is generated, which we can observe from Figure \ref{fgsm boundary along fgsm}, \ref{fgsm boundary along df} intuitively. As shown on Figure \ref{compare accuracy between DF-1}, for \textbf{$\mbox{DF}^{\infty}$-1}, after CO happens, small perturbation along FGSM and $\mbox{DF}^{\infty}$-1 direction is still less effective than the large one.

In this section, we first analyze the geometric properties of classifiers trained by FGSM and $\mbox{DF}^{\infty}$-1 before and after CO separately. And we find FGSM generates a new decision boundary both along FGSM direction and $\mbox{DF}^{\infty}$-1 direction, which makes small perturbation becomes less effective than large perturbation. As for $\mbox{DF}^{\infty}$-1, the marginal between clean input and decision boundary becomes smaller but is still larger than FGSM after CO. And $\mbox{DF}^{\infty}$-1 does not generate a new decision boundary and large perturbation is still more effective than small perturbation. But the perturbation generated by $\mbox{DF}^{\infty}$-1 becomes smaller after CO and thus loses its effectiveness. The geometric analysis demonstrates the decision boundaries after CO and shows why adversary methods used in training lose their effectiveness, but why CO happens still needs more exploration. In next section, we will focus on analyzing factors that cause CO.

\section{Analysis on factors causing catastrophic overfitting}
In the last section, we find geometric properties of the classifier change after CO and make both FGSM and $\mbox{DF}^{\infty}$-1 adversarial methods become ineffective. This explains why the model can not recover by itself once CO happens. The difficulty to bring the model back to robust after CO happens increases the importance of preventing the occurrence of CO. In this section, we will focus on analyzing factors that cause CO. Specifically, we design experiments to examine three probable hypotheses, two of them come from previous literature and one of them proposed by us. And finally, we make a small modification to RS-FGSM which improve its performance.

\subsection{Probable hypotheses on factors causing catastrophic overfitting}
\subsubsection{Hypothesis: large perturbation causes catastrophic overfitting} \label{hypo1}
\textbf{Hypothesis} The length of perturbation is evaluated by $l_2$ norm. Large perturbation causes CO while small perturbation can avoid CO.

\begin{figure}[H]
\begin{center}
\subfloat[]{
  \includegraphics[width=0.53\textwidth]{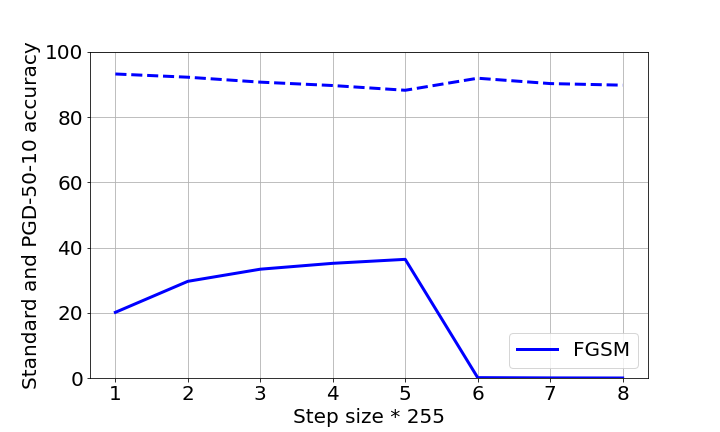} \label{fgsm_step_size}
}
\subfloat[]{
  \includegraphics[width=0.45\textwidth]{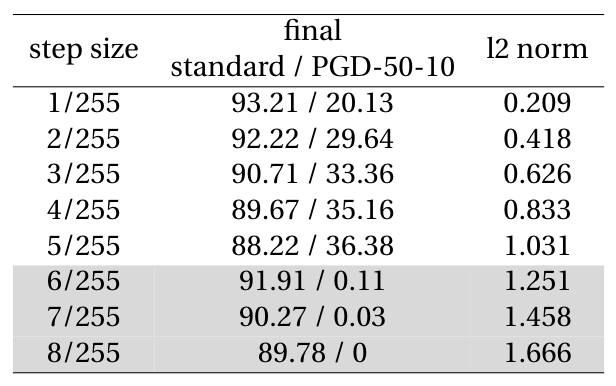} \label{acc_final_fgsm_step_size}
}
\caption{We train PreAct-ResNet18 on CIFAR10 using FGSM with different step size $\alpha$ and fixed $\epsilon=8/255$.}
\label{fgsm_reduce_step_size}
\end{center}
\end{figure}

This hypothesis is first proposed in paper \cite{andriushchenko2020understanding}. The author claims that RS-FGSM helps avoid CO by decreasing the expected $l_2$ norm of  perturbations. And the paper provides theoretical proof that RS-FGSM always has a smaller expected $l_2$ norm compared to vanilla FGSM. We also find another support phenomenon that for a fixed $\epsilon$, reducing the step size $\alpha$ can avoid CO. As shown in Figure \ref{fgsm_reduce_step_size}, we train PreAct-ResNet18 on CIFAR10 using vanilla FGSM. We observe that for the fixed $\epsilon=8/255$, we can avoid catastrophic overfitting when we reduce the step size to $\alpha \leq 5/255$. And smaller step size leads to smaller perturbations. Though reducing the step size $\alpha$ can avoid CO, it will lead to the sub-optimal solution since using a smaller step size is equivalent to training the model with perturbations from a smaller $l_\infty$ ball than the one used during the test.


\begin{equation} \label{Magnified-RS-FGSM}
    \begin{aligned}
    \delta_{FGSM}&=\epsilon \mbox{sign}(\nabla_{x}l(x,y;\theta)) \\
    \delta &\sim \mathcal{U}([-\epsilon, \epsilon]^d) \\
    \delta_{RS-FGSM} &= \textstyle\prod_{[-\epsilon, \epsilon]^d}(\delta +\alpha \mbox{sign}(\nabla_{x}l(x+\delta,y;\theta))) \\
    \delta_{magnified} &= \frac{\|\delta_{FGSM}\|_2}{\|\delta_{RS-FGSM}\|_2} \delta_{rs}
    \end{aligned}
\end{equation}

We design the \textbf{Magnified-RS-FGSM} method to examine this hypothesis. As shown in Equation \ref{Magnified-RS-FGSM}, we first compute $\delta_{FGSM}$ using vanilla FGSM, then we choose a step size $\alpha$ and calculate the perturbation $\delta_{RS-FGSM}$ using RS FGSM, and finally magnify it by $\frac{\|\delta_{FGSM}\|_2}{\|\delta_{RS-FGSM}\|_2}$ to make the $l_2$ norm of this perturbation same as the perturbation calculated by vanilla FGSM $\delta_{FGSM}$.
 
\begin{figure}[H]
\begin{center}
\subfloat[]{
  \includegraphics[width=0.33\textwidth]{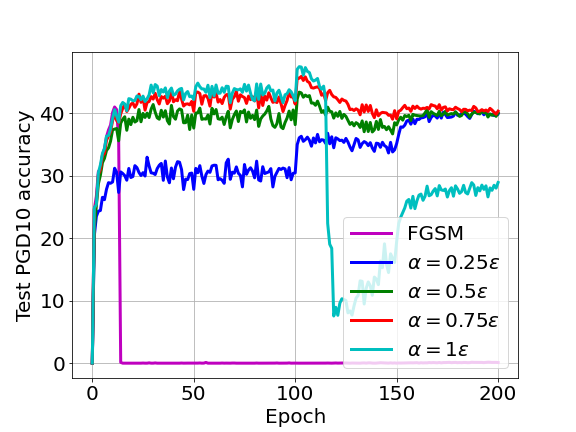} \label{extend_acc}
}
\subfloat[]{
  \includegraphics[width=0.33\textwidth]{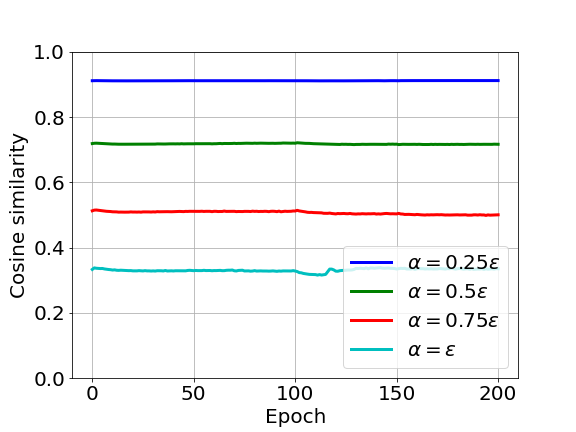} \label{extend_cos_random_delta}
}
\subfloat[]{
  \includegraphics[width=0.33\textwidth]{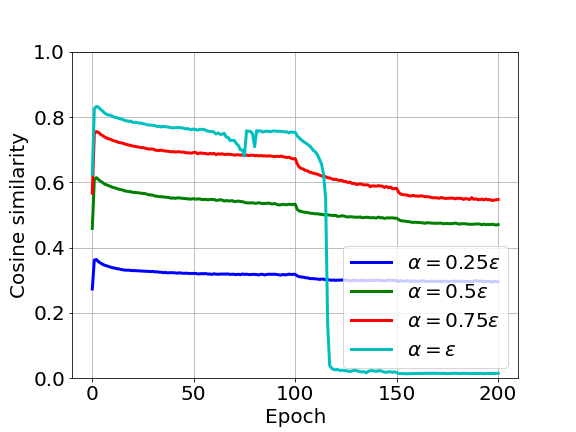} \label{extend_cos_random_grad}
}
\caption{The PreAct-ResNet18 trained by \textbf{Magnified-RS-FGSM} magnified from RS-FGSM with different step size $\alpha$ and fixed $\epsilon=8/255$. (a) Show the test PGD10 accuracy during the training process with vanilla FGSM and Magnified-RS-FGSM magnified from RS-FGSM with different step size $\alpha$. (b)The cosine similarity between $\delta_{RS-FGSM}$ and random initialized $\delta$. (c) The cosine similarity between $\delta_{RS-FGSM}$ and sign of the input gradient $\mbox{sign}(\nabla_{x}l(x,y;\theta))$}
\label{figure:extend}
\end{center}
\end{figure}

As shown in Figure \ref{extend_acc}, when step size $\alpha \leq 0.75 \epsilon$ and $\epsilon=8/255$, we can magnify the perturbation to the same $l_2$ norm as the perturbation calculated by vanilla FGSM without suffering from CO. This experiment result shows that $l_2$ norm of perturbation is not the only factor that decides whether CO happens or not, but the direction of the perturbation is also important.

Thus, we calculate the cosine similarity between perturbation vector $\delta_{RS-FGSM}$ and random initialized vector $\delta$,  and the cosine similarity between perturbation vector $\delta_{RS-FGSM}$ and sign of the input gradient vector $\mbox{sign}(\nabla_{x}l(x,y;\theta))$. The results are shown in Figure \ref{extend_cos_random_delta} \ref{extend_cos_random_grad}, the direction of perturbation vector $\delta_{RS-FGSM}$ computed from smaller step size $\alpha$ is closer to the random initialized direction and farther away from the direction of sign of the input gradient. And we find perturbation computed by smaller step size can be magnified to the same $l_2$ norm as the perturbation calculated by vanilla FGSM while larger step size can not. Based on this phenomenon, we come up with the hypothesis that perturbation computed from smaller step size is closer to the random initialized direction and can be magnified to a larger perturbation without suffering from CO.

\begin{table}[H]
\centering
\begin{adjustbox}{max width=\textwidth}
\begin{tabular}{lcc}
\hline
& \begin{tabular}[c]{@{}c@{}}final\\ standard / PGD-50-10\end{tabular} & \begin{tabular}[c]{@{}c@{}}best\\ standard / PGD-50-10\end{tabular} \\ \hline
\multicolumn{3}{c}{$\epsilon$ = 8 / 255} \\ \hline
$\alpha = 0.25\epsilon $ & 90.03 / 34.23  & 89.88 / 34.05  \\
$\alpha = 0.5\epsilon $ & 87.06 / 36.36  & 86.66 / 40.94  \\
$\alpha = 0.75\epsilon $ & 85.41 / 36.4   & 85.54 / 43.8   \\
\rowcolor[HTML]{D9D9D9} 
$\alpha = \epsilon $    & 90.17 / 10.09  & 84.42 / 45.35  \\
\rowcolor[HTML]{D9D9D9} 
FGSM         & 91.07 / 0      & 66.72 / 40.46  \\ \hline
\multicolumn{3}{c}{$\epsilon$ = 12 / 255} \\ \hline
$\alpha = 0.25\epsilon $& 86.83 / 19.19  & 86.77 / 19.15  \\
$\alpha = 0.5\epsilon $  & 83.18 / 18.96  & 83.11 / 25.02  \\
\rowcolor[HTML]{D9D9D9} 
$\alpha = 0.75\epsilon $ & 87.79 / 0.04   & 70.8 / 30.11   \\
\rowcolor[HTML]{D9D9D9} 
$\alpha = \epsilon $  & 88.55 / 0      & 58.11 / 29.69  \\
\rowcolor[HTML]{D9D9D9} 
FGSM         & 88.35 / 0      & 46.59 / 25.09  \\ \hline
\end{tabular}
\end{adjustbox}
\caption{Show the standard and PGD-50-10 accuracy of the final and best model trained by Magnified-RS-FGSM magnified from RS-FGSM with different step size $\alpha$ under $\epsilon=8/255$ or $\epsilon=12/255$. The gray background of the cell indicates that CO happens under this settings.}
\label{table:extend}
\end{table}

In order to verify our new hypothesis, we run the same experiments with $\epsilon=12/255$ and then evaluate the robustness of the best and final model using PGD-50-10. The result is shown on Table \ref{table:extend}. We observe that when we need to magnify the perturbation to the same $l_2$ norm as vanilla FGSM with $\epsilon=12/255$, $\alpha = 0.25\epsilon$ and $\alpha = 0.5\epsilon$ can still avoid catastrophic overfitting while $\alpha = 0.75\epsilon$ suffers from catastrophic overfitting. This supports the conclusion that perturbation computed from smaller step size can be magnified to a larger perturbation without suffering from CO. Another observation is that all step sizes $\alpha \leq 0.75\epsilon$ can avoid CO when $\epsilon=8/255$. $\alpha = 0.75\epsilon$ has the best robust accuracy of 43.8\% which is 9.75\% better than $\alpha = 0.25\epsilon$, even if they have the same $l_2$ norm as the perturbation calculated by vanilla FGSM. This indicates that under the same perturbation length if the direction of perturbation is closer to the random initialized vector, the robust accuracy will become smaller than the one farther away from the random initialized vector and closer to the sign of the input gradient vector. This means even if by choosing direction we can get the same perturbation length as vanilla FGSM, we still get the sub-optimal solution.

In this subsection, we design the experiments to examine the hypothesis that large perturbation can cause CO. And finally we conclude that not only $l_2$ norm of the perturbation matters but also the direction of the perturbation.

\subsubsection{Hypothesis: perturbation should span the entire model} \label{hypo2}
The paper\cite{wong2020fast} claims that RS-FGSM helps avoid CO by distributing perturbation features into the entire threat model $[-\epsilon, \epsilon]^d$; in other words, each dimension can take value between $[-\epsilon, \epsilon]$. There are two evidences to support this hypothesis. One is that if we choose the step size $\alpha=2\epsilon$ to force each dimension to be \{$-\epsilon$, $\epsilon$\}, RS-FGSM also suffers from catastrophic overfitting when $\epsilon=8/255$. 
\begin{equation} \label{R+FGSM}
    \begin{aligned}
    \delta &\sim \{-\frac{\epsilon}{2}, \frac{\epsilon}{2}\}^d\\
    \delta_{R+FGSM} &= \textstyle\prod_{[-\epsilon, \epsilon]^d}(\delta +\frac{\epsilon}{2} \mbox{sign}(\nabla_{x}l(x+\delta,y;\theta)))\\
    \end{aligned}
\end{equation}
The other is that R+FGSM from \cite{tramèr2020ensemble}, as shown in Equation \ref{R+FGSM}, cannot train the robust model on MNIST dataset as shown on \cite{wong2020fast} appendix-A while RS-FGSM is capable. The difference between R+FGSM and RS-FGSM is that R+FGSM can only generate perturbation with features on the $l_\infty$ ball $\{-\epsilon, 0, \epsilon\}^d$ while RS-FGSM can take value between $[-\epsilon, \epsilon]$.

\begin{equation} \label{Boundary-RS-FGSM}
    \begin{aligned}
    \delta &\sim \{-\epsilon, \epsilon\}^d \\
    \delta_{Boundary-RS-FGSM} &= \textstyle\prod_{[-\epsilon, \epsilon]^d}(\delta +\alpha \mbox{sign}(\nabla_{x}l(x+\delta,y;\theta)))\\
    \end{aligned}
\end{equation}

we design the following \textbf{Boundary-RS-FGSM} method to examine the hypothesis that perturbation should span the entire model; in other words, each dimension of the perturbation can take value between $[-\epsilon, \epsilon]$.  As shown in Equation \ref{Boundary-RS-FGSM}, when we perform the random initialization, instead of taking values uniformly distributed from $[-\epsilon, \epsilon]^d$, we randomly choose either $-\epsilon$ or $\epsilon$ for each dimension of the perturbation. In this way, if the step size $\alpha=\epsilon$, the final perturbation generated will have features on $\{-\epsilon, 0, \epsilon\}$; if the step size $\alpha=1.5\epsilon$, the final perturbation generated will have features on $\{-\epsilon, -0.5\epsilon, 0.5\epsilon, \epsilon\}$. They are discrete values and not span inside the $l_\infty$ ball.

\begin{figure}[H]
 \begin{center}
  \includegraphics[width=0.6\textwidth]{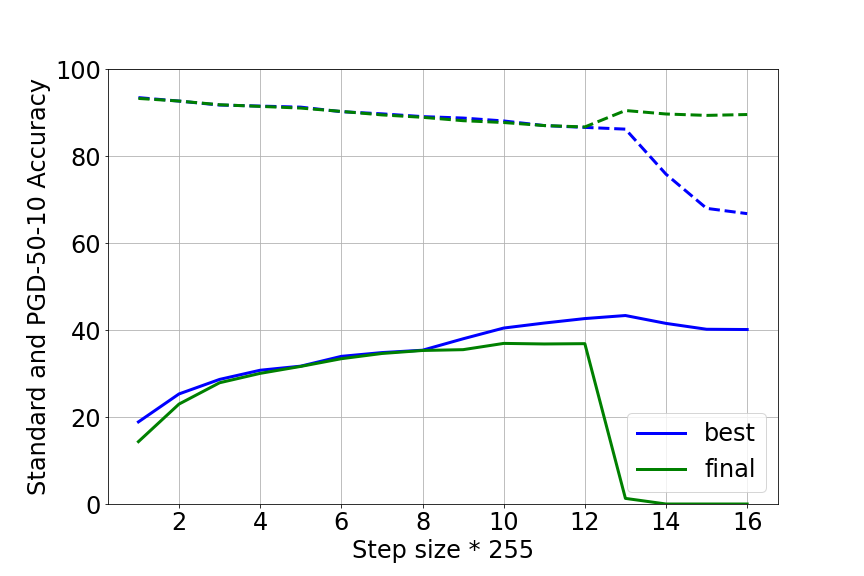}
  \caption{Standard (dashed line) and PGD-50-10 accuracy (solid line) of \textbf{Boundary-RS-FGSM} over different step size $\alpha$ with fixed $\epsilon=8/255$. We evaluate both the best model and the final model.}
  \label{figure:boundary_rs_fgsm}
 \end{center}
\end{figure}

As shown on the Figure \ref{figure:boundary_rs_fgsm}, we train the PreAct-ResNet18 model using \textbf{Boundary-RS-FGSM} adversarial training with fixed $\epsilon=8/255$ and different step size $\alpha$ on CIFAR10. We observe that catastrophic overfitting happens when step-size $\alpha \geq 1.5\epsilon$. So when $\alpha=1.5\epsilon$, we can get the best robust accuracy without suffering from CO. 

\begin{table}[H]
\centering
\begin{adjustbox}{max width=\textwidth}
\begin{tabular}{cc}
\hline
method            & \begin{tabular}[c]{@{}c@{}}best\\ standard / PGD-50-10\end{tabular} \\ \hline
\rowcolor[HTML]{D9D9D9} 
FGSM              & 66.72 / 40.46                                                    \\
RS-FGSM & 86.77 / 42.69                                                    \\
Boundary-RS-FGSM  & 87.03 / 42.72                                        \\ \hline
\end{tabular}
\end{adjustbox}
\caption{Comparison of the standard and PGD-50-10 performance on CIFAR10 with $\epsilon=8/255$ between vanilla FGSM, RS-FGSM and Boundary-RS-FGSM. We use the step size $\alpha=\epsilon$ for RS-FGSM, and $\alpha=1.5\epsilon$ for Boundary-RS-FGSM.}
\label{table:boundary_rs_fgsm}
\end{table}

We also compare the standard and PGD-50-10 performance between RS-FGSM and Boundary-RS-FGSM. As shown in Table \ref{table:boundary_rs_fgsm}, they can achieve almost the same performance on CIFAR10 with $\epsilon=8/255$. The experiment results invalidate this hypothesis that each dimension of the perturbation should span in the entire threat model.

We also apply the Boundary-RS-FGSM to train the robust model on MNIST dataset. We use the same setting as the paper \cite{wong2020fast}. And based on the results on the Table \ref{minst_fgsm_boundary}, Boundary-RS-FGSM also achieves almost the same robust accuracy as RS-FGSM.

\begin{table}[H]
\centering
\begin{adjustbox}{max width=\textwidth}
\begin{tabular}{lll}
Method                         & Step size   & PGD-50-10 \\ \hline
R+FGSM    & $0.5\epsilon$       & 31.83$\pm$25.36\%    \\
RS-FGSM & $\epsilon$         & 86.11$\pm$1.4\%     \\
Boundary-RS-FGSM                  & $\epsilon$      & 80.64$\pm$2.22\%      \\
Boundary-RS-FGSM                  & $1.5\epsilon$   & 85.44$\pm$2.1\%       \\ \hline
\end{tabular}
\end{adjustbox}
\caption{Show the performance of R+FGSM \cite{tramèr2020ensemble}, RS-FGSM \cite{wong2020fast} and Boundary-RS-FGSM on MNIST dataset over 5 random seeds.}
\label{minst_fgsm_boundary}
\end{table}

\subsubsection{Hypothesis: large diversity of perturbations can avoid catastrophic overfitting} \label{hypo3}

\begin{equation}\label{equ: diversity}
    Diversity = 1 - cos(\delta_a, \delta_b)
\end{equation}

We define the diversity metric of perturbation as Equation \ref{equ: diversity}. We compute perturbations twice using the same input and model. $\delta_a$  is the first one and $\delta_b$ is the second one.

\begin{equation} \label{Diff-RS-FGSM}
    \begin{aligned}
    \delta_{1}, \delta_{2} &\sim \mathcal{U}([-\epsilon, \epsilon]^d) \\
    \delta &= (1-t)\delta_{1} + t\delta_{2}, \mbox{ where } t\in [0,1] \\
    \delta_{Diff-RS-FGSM} &= \textstyle\prod_{[-\epsilon, \epsilon]^d}(\delta_{1} + \alpha \mbox{sign}(\nabla_{x}l(x+\delta,y;\theta)))\\
    \end{aligned}
\end{equation}

As for the same input and model, FGSM perturabtions are always the same, it has zero diversity. But for the RS-FGSM, since we have random initialization step, $\delta_a$ and $\delta_b$ will be different and has positive diversity. Thus, we can have the hypothesis that large diversity of the perturbation can avoid CO. In order to examine this hypothesis, we design the following \textbf{Diff-RS-FGSM} method, as shown in Equation \ref{Diff-RS-FGSM}.

\begin{figure}[H]
\begin{center}
\subfloat[]{
  \includegraphics[width=0.45\textwidth]{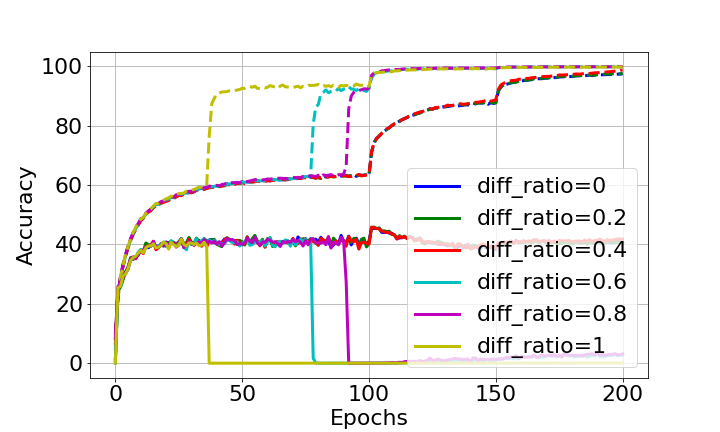} \label{diff_rs_acc}
}
\subfloat[]{
  \includegraphics[width=0.45\textwidth]{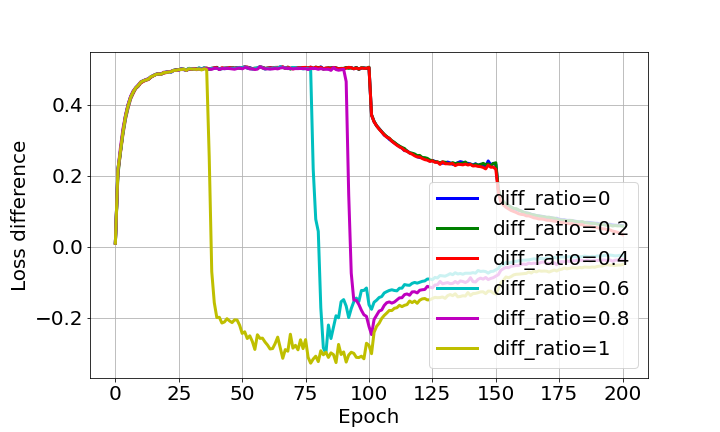} \label{diff_rs_loss}
}
\caption{Train the PreAct-ResNet18 model using \textbf{Diff-RS-FGSM} over different diff-ratio $t$ on CIFAR10 with fixed $\epsilon=8/255$ and step size $\alpha=\epsilon$. (a) Show Diff-RS-FGSM accuracy on training datasert (dashed line) and PGD-10 accuracy on testing dataset (solid line) during the training process. (b) Show the loss difference between train\_Diff-RS-FGSM\_loss and train\_standard\_loss.}
\label{figure: diff rs}
\end{center}
\end{figure}

\begin{figure}[H]
 \begin{center}
  \includegraphics[width=0.6\textwidth]{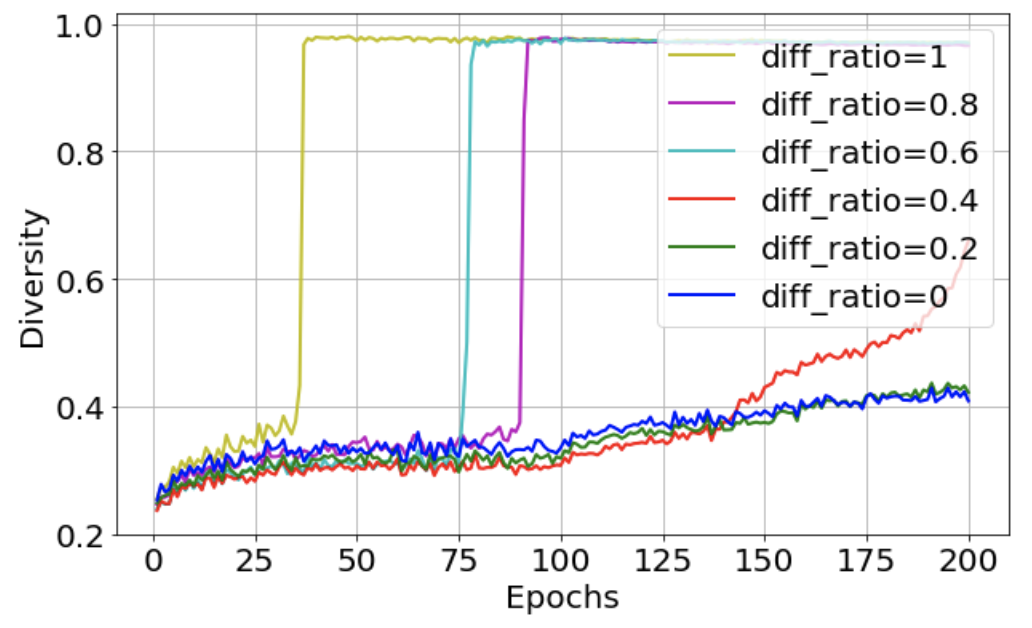}
  \caption{Diversity of the Diff-RS-FGSM under different diff-ratio $t$}
  \label{diversity}
 \end{center}
\end{figure}

When diff-ratio $t=0$, Diff-RS-FGSM is the same as RS-FGSM and does not suffer from catastrophic overfitting. As shown in Figure \ref{figure: diff rs} the larger diff-ratio $t$, the earlier  catastrophic overfitting happens. Another phenomenon we notice is that before catastrophic overfitting happens, Diff-RS-FGSM training with diff-ratio $t$ has almost the same accuracy and loss difference between perturbed points and clean points.  

As shown in Figure \ref{diversity}, Diff-RS-FGSM with different diff-ratio $t$ has almost the same diversity before CO. But the larger the diff-ratio $t$ and the earlier the catastrophic overfitting happens. And once CO happens, the diversity will increased to almost 1. So we reject this hypothesis by saying that large diversity cannot guarantee to avoid CO.

\subsection{Further improvement on RS-FGSM methods}
In the previous section, we empirically analyze three hypotheses on factors causing catastrophic overfitting. For each hypothesis we comp up with counter experiments to show that none of these hypothesis can fully explain why CO happens. But these analyses do shed light on understanding why catastrophic overfitting happens. Based on the Section \ref{hypo1}, not only $l_2$ norm will affect CO, but also the direction of perturbation. If the direction of perturbation is closer to the random initialized vector, the robust accuracy will become smaller than the one far away from the random initialized vector and closer to the sign of the input gradient vector. So we try the RS- FGSM without projecting back to $l_\infty$-ball to follow the sign of the input gradient vector as much as possible. We name the RS-FGSM and Boundary-RS-FGSM without projecting back to \textbf{RS-FGSM-wo-Proj} and \textbf{Boundary-RS-FGSM-wo-Proj} separately.

\begin{figure}[H]
\begin{center}
\subfloat[Final]{
  \includegraphics[width=0.5\textwidth]{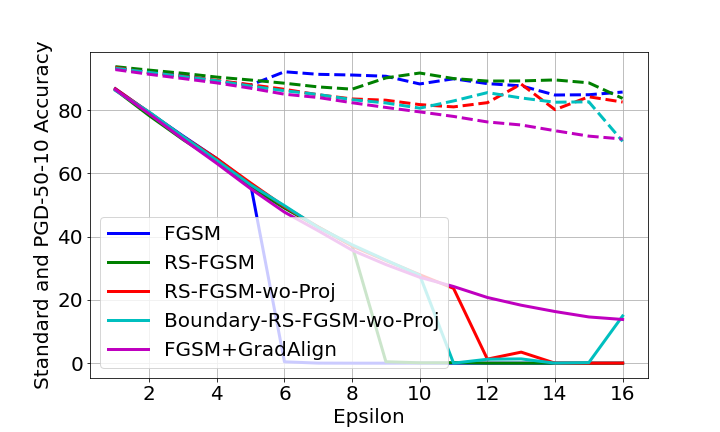} \label{final_compare}
}
\subfloat[Best]{
  \includegraphics[width=0.5\textwidth]{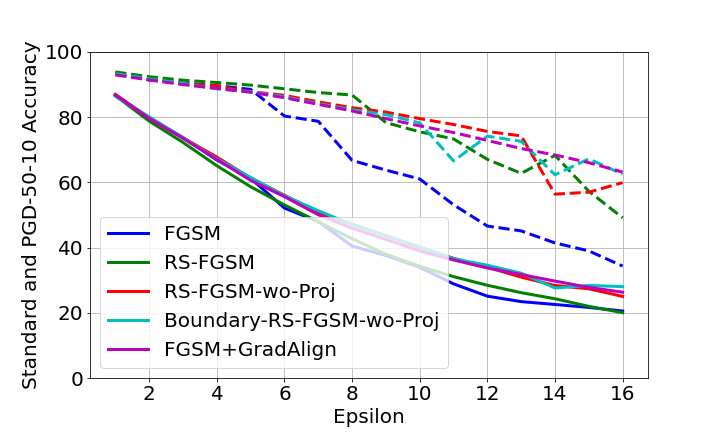} \label{best_compare}
}
\caption{Standard accuracy (dashed line) and PGD-50-10 accuracy (solid line) from different FGSM methods with different $\epsilon$. (a) Results of the final model showing whether CO happens or not. (b) Results of the best model selected by PGD-10 during the training process.}
\label{figure: wo_proj}
\end{center}
\end{figure}

As shown in Figure \ref{figure: wo_proj}, we observe that RS-FGSM-wo-proj further extends the $\epsilon$ above which CO happens. RS-FGSM extends the working regime of $\epsilon$ from 5/255 to 8/255 compared to vanilla FGSM. And RS-FGSM-wo-Proj further extends the $\epsilon$ to 11/255. Another benefit is that given a fixed $\epsilon$ with which both RS-FGSM and RS-FGSM-wo-Proj do not have catastrophic overfitting, RS-FGSM-wo-Proj still has better robust PGD-50-10 accuracy and even comparable to FGSM+GradAlign\cite{andriushchenko2020understanding}.

As shown on Table \ref{table: wo_proj}, when $\epsilon=8/255$, Both RS-FGSM-wo-Proj and Boundary-RS-FGSM-wo-proj have comparable performance as FGSM+GradAlign with better computational efficiency. When $\epsilon=16/255$, the average performance of RS-FGSM-wo-Proj and Boundary-RS-FGSM-wo-proj is comparable to FGSM+GradAlign, but has larger standard deviation.

In this section, we focus on analyzing factors that cause CO. Specifically, we design experiments to examine three probable hypotheses. Although finally, we find none of them can fully explain why CO happens but the analysis process and results of designed experiments do improve the understanding toward catastrophic overfitting. Then in the second section, we make a small modification to RS-FGSM by not projecting back to $l_\infty$ thread model and gain further improvements on RS-FGSM.

\begin{table}[H]
\centering
\begin{adjustbox}{max width=\textwidth}
\begin{tabular}{lccl}
\hline
\multicolumn{1}{c}{\multirow{2}{*}{Method}} & \multicolumn{2}{c}{Accuracy}                              & \multicolumn{1}{c}{\multirow{2}{*}{Attack}} \\
\multicolumn{1}{c}{}                        & \multicolumn{1}{l}{Standard} & \multicolumn{1}{l}{Robust} & \multicolumn{1}{c}{}                        \\ \hline
\multicolumn{4}{c}{$\epsilon$=8 / 255, step\_size = $\epsilon$}                                                                    \\
FGSM                                        &  69.21$\pm$2.18\%    &  41.19$\pm$0.38\% & PGD-50-10    \\
RS-FGSM                                     &  86.35$\pm$0.34\%    &  43.57$\pm$0.30\% & PGD-50-10  \\
RS-FGSM-wo-proj                            &  82.66$\pm$0.56\%    &  47.56$\pm$0.37\% & PGD-50-10  \\
Boundary-RS-FGSM-wo-proj                   &  82.29$\pm$0.46\%    &  47.65$\pm$0.52\% & PGD-50-10  \\
FGSM + GradAlign                 &  81.34$\pm$0.45\%    &  46.63$\pm$0.52\% & PGD-50-10  \\
PGD-10($\alpha=2\epsilon/10$)       &  82.69$\pm$0.62\%    &  50.14$\pm$0.64\%& PGD-50-10  \\ \hline
\multicolumn{4}{c}{$\epsilon$=16 / 255, step\_size = $\epsilon$}                                                     \\
FGSM                                        &  34.42$\pm$2.61\% &  20.41$\pm$0.95\%& PGD-50-10 \\
RS-FGSM                                     &  51.87$\pm$3.29\% &  21.28$\pm$0.92\%& PGD-50-10 \\
RS-FGSM-wo-proj                            &  57.89$\pm$5.82\% &  25.63$\pm$0.38\%& PGD-50-10 \\
Boundary-RS-FGSM-wo-proj                   &  57.32$\pm$8.06\% &  26.32$\pm$1.42\% & PGD-50-10 \\
FGSM + GradAlign                         &  63.20$\pm$1.03\% &  26.04$\pm$0.66\% & PGD-50-10 \\
PGD-10($\alpha=2\epsilon/10$)                  &  65.95$\pm$1.40\% &  32.86$\pm$0.50\% & PGD-50-10 \\ \hline
\end{tabular}
\end{adjustbox}
\caption{Robustness and accuracy of different robust training methods on CIFAR-10. And the results are shown here with the standard deviation and averaged over 5 random seeds used for training. We report the results by selecting the best test robust accuracy. We reproduce the FGSM+GradAlign\cite{andriushchenko2020understanding} and PGD-10 accuracy} 
\label{table: wo_proj}
\end{table}

\section{Conclusion}
We observe that catastrophic overfitting is a general phenomenon in adversarial training, which not only occurs in FGSM and RS-FGSM adversarial training but also occurs in $\mbox{DF}^{\infty}$-1 adversarial training. Thus, it is important to understand catastrophic overfitting and prevent it from happening in order to train the model using weak adversaries and gain robustness against strong adversaries. 

We find after catastrophic overfitting happens, the marginal between clean inputs and decision boundary decreases for both FGSM and $\mbox{DF}^{\infty}$-1 trained model. But the mechanisms why FGSM and $\mbox{DF}^{\infty}$-1 adversary loses its effectiveness are different. For FGSM, a new decision boundary is generated along the direction of perturbation and makes the small perturbation more effective than the large perturbation. However, for $\mbox{DF}^{\infty}$-1, there is no decision boundary generated along the direction of perturbation. And large perturbation is still more effective than small perturbation. But the perturbations generated by $\mbox{DF}^{\infty}$-1 becomes smaller after catastrophic overfitting and thus loses their effectiveness. 

As for factors that cause catastrophic overfitting, we find not only $l_2$ norm of the perturbation is important but also the direction of the perturbation. Besides, the perturbation is not necessary to span in the entire model and large diversity is not enough to avoid catastrophic overfitting. Finally, we propose a modification to RS-FGSM. After calculating the perturbation, we do not project it back to $l_{\infty}$-ball to make this perturbation follow the direction of the sign of the input gradient vector as much as possible. This modification permits us to use larger values of $\epsilon$ and improves the robust accuracy compared to RS-FGSM under the same $\epsilon$.

\textbf{Future work}

To resolve the catastrophic overfitting problem in adversarial training, there are two main research directions. One is to understand why CO happens and prevent it before it happens, and the other one is to analyze the properties after CO happens and recover the training from CO. 

In this project, the geometric properties after catastrophic overfitting happens are well studied for both FGSM and $\mbox{DF}^{\infty}$-1. The remaining question here is why FGSM and $\mbox{DF}^{\infty}$-1 show totally different geometric properties after CO happens. And this question needs further exploration.

As for why catastrophic overfitting happens, we design experiments to analyze three hypotheses on potential factors causing CO, but find that none of them can fully explain why CO happens. So we need to put more efforts to study the main factors that cause CO. Below are some potential research questions:
\begin{itemize}
    \item Explore the relationship between the direction of the perturbation and the maximum length of the perturbation which does not cause CO. This can be studied both theoretically and empirically.
    \item In RS-FGSM, we use this equation $\textstyle\prod_{[-\epsilon,\epsilon]^d}(\delta+\alpha\mbox{sign}(\nabla_{x}l(x+\delta,y;\theta)))$ to calculate perturbations. The random initialized $\delta$ shows in two places. One is $\nabla_{x}l(x+\underline{\bm{\delta}},y;\theta)$ to add randomness into the gradient direction inside gradient, and the other one is $\underline{\bm{\delta}}+\alpha \mbox{sign}(\nabla_{x}l(x+\delta,y;\theta))$ to add randomness into the length of each perturbation dimension  outside gradient. We find removing either of them $\delta$ will cause the RS-FGSM training to fail. Thus, we can study the usage of $\delta$ in these two places and the roles they play in RS-FGSM to mitigate CO.
\end{itemize}

\phantomsection
\addcontentsline{toc}{section}{References}
\printbibliography

\end{document}